\documentclass{article}
\PassOptionsToPackage{numbers, compress}{natbib}
\usepackage{pifont}
\usepackage{pdflscape} 
\usepackage{booktabs}   
\usepackage{placeins}
\usepackage{wrapfig}
\usepackage[skip=3pt]{caption}
\usepackage{threeparttable}
\usepackage[main, final]{neurips_2026}
\usepackage{multirow,graphicx,booktabs}

\usepackage[utf8]{inputenc}
\usepackage[T1]{fontenc}
\usepackage{amsthm}

\usepackage{enumitem}
\usepackage{multirow}

\usepackage{newunicodechar}

\usepackage[utf8]{inputenc} 
\usepackage[T1]{fontenc}    
\usepackage{hyperref}       
\usepackage{url}            
\usepackage{booktabs}       
\usepackage{amsfonts}       
\usepackage{nicefrac}       
\usepackage{microtype}      
\usepackage{xcolor}         
\usepackage{amsmath}
\usepackage{amssymb}
\usepackage{booktabs}
\usepackage{graphicx}
\usepackage{xcolor}
\usepackage{booktabs}
\usepackage{graphicx}
\usepackage{xcolor}
\usepackage{amssymb}
\usepackage{array}
\usepackage{tabularx}
\usepackage{algorithm}
\usepackage{algpseudocode}

\usepackage{setspace}

\usepackage{booktabs}
\usepackage{multirow}
\usepackage{graphicx}                 
\usepackage{tabularx}
\usepackage[table]{xcolor}
\usepackage[nointegrals]{wasysym}     

\newcommand{\cmark}{\ding{51}}      
\newcommand{\xmark}{\ding{55}}      
\newcommand{\pmark}{\ensuremath{\sim}}  
\definecolor{oursrow}{gray}{0.92}
\definecolor{amber}{RGB}{204,153,0}

\newcommand{\bestall}[1]{\cellcolor{red!18}\textbf{#1}}
\newcommand{\secondall}[1]{\cellcolor{yellow!35}\textbf{#1}}
\newcommand{\bestint}[1]{\cellcolor{blue!15}\textbf{#1}}
\newcommand{\capbox}[2]{{\setlength{\fboxsep}{1pt}\colorbox{#1}{\textbf{#2}}}}
\newcommand{\capbest}[1]{\capbox{red!18}{#1}}
\newcommand{\capsecond}[1]{\capbox{yellow!35}{#1}}
\newcommand{\capint}[1]{\capbox{blue!15}{#1}}

\definecolor{lightgrayrow}{gray}{0.95}
\newunicodechar{◐}{\CircleHalf}
\theoremstyle{remark}

\definecolor{oursrow}{gray}{0.92}
\newcolumntype{C}{>{\centering\arraybackslash}X}
\algrenewcommand\alglinenumber[1]{\scriptsize #1:}
\newcommand{\ct}{\texttt{ConceptTime}}

\algrenewcommand\alglinenumber[1]{\scriptsize #1:}
\algrenewcommand\algorithmicindent{0.9em}

\providecommand{\cmark}{}\renewcommand{\cmark}{\CIRCLE}
\providecommand{\pmark}{}\renewcommand{\pmark}{\LEFTcircle}
\providecommand{\xmark}{}\renewcommand{\xmark}{\Circle}

\title{Predictive Surprise as Self-Grounding Concept Bottleneck for Interpretable Time Series}

\author{%
Sachith Abeywickrama\textsuperscript{1,2}
\quad
Emadeldeen Eldele\textsuperscript{3}
\quad
Min Wu\textsuperscript{2}\thanks{Corresponding author. \texttt{wumin@a-star.edu.sg}}
\quad
Xiaoli Li\textsuperscript{4}
\quad
Chau Yuen\textsuperscript{1}
\\[1.5mm]
\textsuperscript{1}Nanyang Technological University, Singapore
\\
\textsuperscript{2}Institute of Advanced Intelligence and Computing (IAIC), A*STAR, Singapore
\\
\textsuperscript{3}Department of Computer Science, Khalifa University, UAE
\\
\textsuperscript{4}Singapore University of Technology and Design, Singapore
}

\begin{document}

\maketitle
\begin{abstract}
Time-series classifiers deployed in risk-sensitive domains must be (1) accurate, (2) interpretable, and (3) able to abstain when evidence is weak.
Existing methods do not deliver all 3 at once, and the obstacle is structural, rather than logistical.
Meaningful temporal patterns cannot be defined without knowing segment boundaries, yet meaningful boundaries cannot be placed without knowing which patterns to expect.
Existing approaches break this circularity by using fixed windows or expert-annotated vocabularies, thereby sacrificing at least one of the 3 properties.
We introduce \texttt{ConceptTime}, which closes the loop through a single self-supervised signal.
A frozen probabilistic forecaster predicts a distribution over the next observation at every timestep of the sequence.
We call the mismatch between its prediction and the realized signal the \emph{predictive surprise}.
This signal both locates segment boundaries and characterizes each resulting segment through the forecaster.
Segments are thus grounded by \emph{how they behaved} relative to expectation, not by \emph{what they look like}.
These summaries cluster into a vocabulary of dynamical regimes that we call \textit{concepts}.
A lightweight head classifies from the concept sequence, and the distance to the nearest concept yields reliability by design.
Across 17 public benchmarks, \texttt{ConceptTime} is competitive with state-of-the-art black-box accuracy, beats every interpretable baseline, and retains near-98\% accuracy on Epilepsy with 1\% of labels.
It further remains competitive with softmax, prototype-distance, and SHAP/TimeX/TimeX++ on OOD detection and deletion faithfulness, without human annotations, language model calls, or domain experts. 
Code is available at: \url{https://github.com/Sachithx/ConceptTime.git}
\end{abstract}

\section{Introduction}
\label{sec:intro}
 
\begin{figure}
    \centering
    \includegraphics[width=1\linewidth]{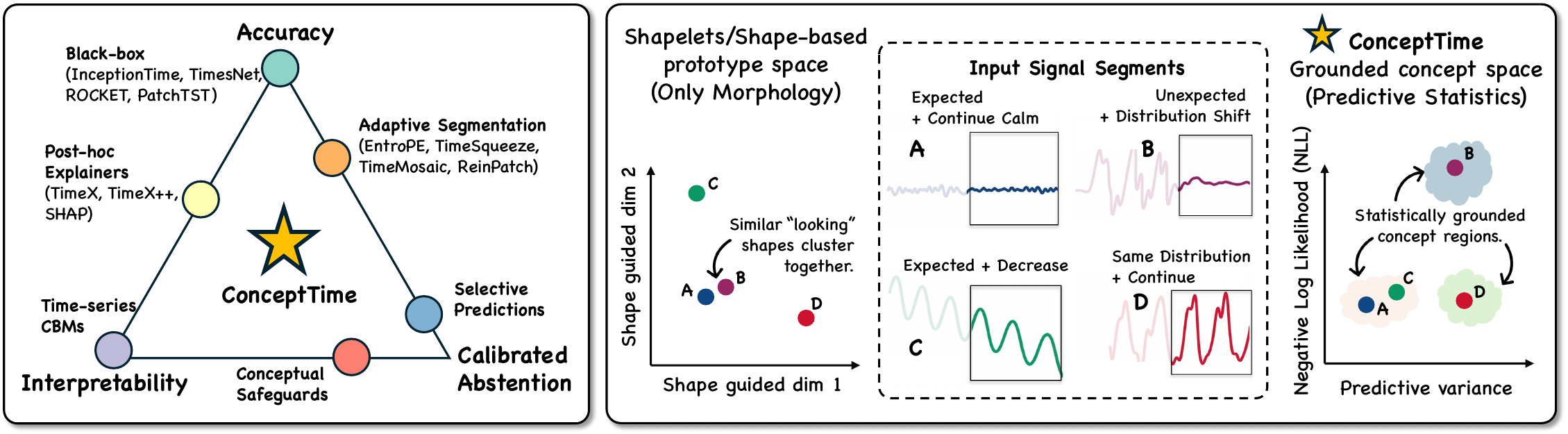}
    \caption{\textbf{\texttt{ConceptTime} targets accuracy, interpretability, and calibrated abstention through a single self-supervised primitive.}
    \emph{Left}, existing methods emphasize different goals.
    \emph{Right}, 4 segments (A-D) and their preceding context. Raw-shape space conflates A and B (both end calm) and splits C from D. Predictive-geometry space, spanned by predictive variance and NLL, collapses A, C, and D as expectation is satisfied and isolates B as a \textit{surprise}. Concepts grounded in deviation from prediction, therefore, distinguish dynamical regimes that shape-based vocabularies cannot.}
    \label{fig:teaser}
\end{figure}

Time-series classifiers deployed in healthcare~\citep{islam2020development, liu2024semi}, wearable sensing~\citep{heikenfeld2018wearable}, environmental monitoring~\citep{ maijala2018environmental}, and industrial telemetry~\citep{ponomarev2015industrial, wickramasinghe2025smart} face 3 simultaneous demands:
(1) predictions must be accurate enough to compete with end-to-end deep models,
(2) they must be interpretable so that users can inspect which temporal patterns drove a decision,
and (3) they must support calibrated abstention, so that the system defers when its evidence is weak.
However, existing methods cannot satisfy all 3 requirements simultaneously (Figure \ref{fig:teaser}).

Black-box classifiers target higher downstream accuracy; post hoc explainable methods describe a model's output without constraining how it was produced, while interpretable-by-design methods come closest, but they fail to address the cost of concept annotation, which is rare for sequential data and expensive to obtain at scale.
The deeper obstacle is structural and unique to sequential domains.
Concept Bottleneck Models (CBMs) for time series face a chicken-and-egg problem:
meaningful temporal concepts cannot be defined without knowing where one segment ends and another begins,
yet meaningful segment boundaries cannot be placed without knowing what concepts to look for.
Every prior method sidesteps this circularity by decoupling the 2 stages.
They first segment with fixed windows or threshold-based change detection, then learn concepts from the resulting chunks.
Each gives up one of the 3 properties as a result.
(a) Fixed-window CBMs~\citep{wu2022learning, sprang2025enforcing, stelle2025} achieve interpretability but require predefined vocabularies and lose calibrated abstention.
(b) Adaptive-patching methods~\citep{abeywickrama2025entrope, pagnoni2025blt} segment well but feed patches into standard encoder-classifier pipelines, exposing no concepts at all.
(c) Vector-quantized models~\citep{talukder2024totem, wen2024abstracted, ansari2024chronos} produce discrete tokens, but the tokens are latent codes without inherent semantic structure.

The same self-supervised signal that locates segment boundaries can also describe what each resulting segment looks like.
A frozen probabilistic forecaster predicts a distribution over the next observation at each time step.
The mismatch between this prediction and the realized signal, which we call \emph{predictive surprise}, is the natural place to draw a boundary, since high mismatch indicates a regime change.
The same forecaster also produces predictive statistics that describe how each segment behaved relative to expectation.
A single frozen model can therefore close the segmentation-concept loop, and no prior method exploits this dual use.

We introduce \texttt{ConceptTime}, which operationalizes this opportunity.
An Autoregressive Gaussian Density Model (ARGDM) is trained once on raw multivariate signals via next-step density estimation.
At each time step, it predicts a Gaussian distribution over the next observation conditioned on the history.
The negative log-likelihood of the realized signal under that prediction is our \emph{predictive surprise}.
This single quantity locates segment boundaries via length-constrained dynamic programming, placing exactly $K$ patches at the moments when the signal most strongly violates prior expectations.
The same model characterizes each resulting patch by its predictive statistics, yielding contextual fingerprints that describe how a segment behaves relative to expectations rather than what it looks like.
Patches that violate prior expectations in similar ways receive similar fingerprints, even when their raw waveforms differ visibly.
The path from raw signal to concept activation is fully deterministic.
Signatures are statistics of the frozen model; prototypes are cluster centroids in signature space; and patch-to-concept assignment is a nearest-prototype lookup.
A lightweight classifier is the only label-supervised component.
The same prototype distances yield a built-in fidelity score that asks whether the input is even \emph{describable} in the concept space rather than whether the classifier is \emph{confident}. 

Across 17 benchmarks spanning UEA, human-activity, biomedical, and fault-diagnosis tasks, \texttt{ConceptTime} remains competitive with black-box accuracy, beats every interpretable baseline, and retains 98\% accuracy on Epilepsy at 1\% labels. Its fidelity score is also competitive with softmax and prototype methods on OOD detection and SHAP~\citep{shap}, TimeX~\citep{queen2023encoding}, and TimeX++~\citep{liu2024timex} on deletion faithfulness, all without human annotations, language-model calls, or domain experts. Further, we demonstrate \texttt{ConceptTime}'s cross-domain transferability of the learned concept under a limited target label budget. Our main contributions can be summarized as follows:
\begin{enumerate}[leftmargin=2em, itemsep=2pt]
    \item To our knowledge, we introduce the first concept-bottleneck framework that resolves the segmentation-concept circularity through a single self-supervised primitive, eliminating the need for fixed windows, expert annotations, and language-model supervision.
    \item We provide a fully non-parametric concept-grounding pipeline whose path from raw signal to concept activation is deterministic and inspectable, with all task-specific learning confined to a lightweight classifier on the concept tensor.
    \item We identify prototype distance as a built-in fidelity signal and propose a prototype-distance abstention mechanism for time-series concept bottlenecks.
\end{enumerate}

\section{Related Work}
\label{sec:related}

\paragraph{Concept bottleneck models.}
Concept bottleneck models (CBMs) route predictions through intermediate concepts that users can inspect and intervene on~\citep{koh2020concept}.
The original formulation requires concept annotations. Later work reduces this burden through label-free concept discovery, including approaches that draw on vision-language models~\citep{oikarinen2023labelfree, yang2023language, srivastava2024vlg, sammani2025clipfree}. Other work studies robustness, counterfactual explanations, and concept locality~\citep{margeloiu2021concept, sawada2022concept, dominici2025counterfactual, jeon2025locality}.
Time-series CBMs use predefined statistical summaries~\citep{wu2022learning}, CKA-aligned concepts for forecasting~\citep{sprang2025enforcing}, or Signal Temporal Logic predicates~\citep{stelle2025}.
These approaches leave open how to discover both a temporal concept and its boundaries when its duration varies and its behavior is difficult to name in advance.

\paragraph{Adaptive patching and discrete tokenization.}
The Byte Latent Transformer (BLT) places boundaries at peaks in next-token entropy~\citep{pagnoni2025blt}. EntroPE adapts this idea to time series by thresholding categorical entropy~\citep{abeywickrama2025entrope}. Related work learns boundaries through reinforcement learning~\citep{wu2026dynamic} or uses content-aware compression~\citep{ankireddy2026timesqueeze}.
A complementary line of work represents time series with discrete tokens, using learned codebooks in TOTEM and VQShape~\citep{talukder2024totem, wen2024abstracted} or value quantization in Chronos~\citep{ansari2024chronos}.
These methods address where to divide a sequence or how to encode it. A further question is whether the predictive signal that selects a boundary can also describe the behavior of the resulting segment.

\paragraph{Prototype, shape-based, and post-hoc methods.}
Prototype and shapelet classifiers explain predictions through representative subsequences and local shape patterns~\citep{ye2009time, ming2019prosenet, grabocka2014learning, ferfoglia2024ecats, wen2025shedding, liu2026unishape}.
Shape alone can miss a distinction that matters for time series. A flat segment may be expected after a quiet interval and surprising after a spike, even when the 2 segments look identical.
Post-hoc explainers identify influential regions through saliency or perturbation~\citep{queen2023encoding, liu2024timex, crabbe2021explaining, huang2026shapex}. Their explanations describe an existing classifier's behavior, while a concept bottleneck requires the classifier to base its prediction on the concepts it exposes.

\paragraph{Selective prediction and concept-level uncertainty.}
Common signals for selective prediction and shift detection include softmax confidence~\citep{geifman2019selective}, ensemble disagreement~\citep{lakshminarayanan2017simple}, and distance-based OOD scores~\citep{lee2018simple, sun2022out, lu2024learning, lu2024diversify}.
In vision, concept-based models support abstention through uncertainty heads, calibration networks, or stochastic concept layers~\citep{joren2024classification, lockhart2022learn, vandenhirtz2024stochastic, kim2023probcbm, schrodi2025selective}.
For time series, a complementary question is whether the learned vocabulary itself can indicate how well it covers a new input.

\paragraph{Positioning of \texttt{ConceptTime}.}
\texttt{ConceptTime} uses one frozen Gaussian density model for both segmentation and concept discovery. Predictive surprise guides patch boundaries, and each patch is summarized by how its observations compare with the model's predictions.
Clustering these summaries yields concept prototypes without concept annotations or linguistic priors. Patches receive soft assignments based on prototype distance, which also supplies a fidelity score for abstention without an additional confidence model.
The UCI-HAR ablation illustrates why the choice of features matters. With the same adaptive patches and vocabulary size, predictive moments achieve 87.7\% accuracy versus 42.7\% for raw-signal morphology, a 45.0-point difference (Section~\ref{sec:ablations}).

\section{ConceptTime}
\label{sec:method}

\begin{figure}
    \centering
    \includegraphics[width=1\linewidth]{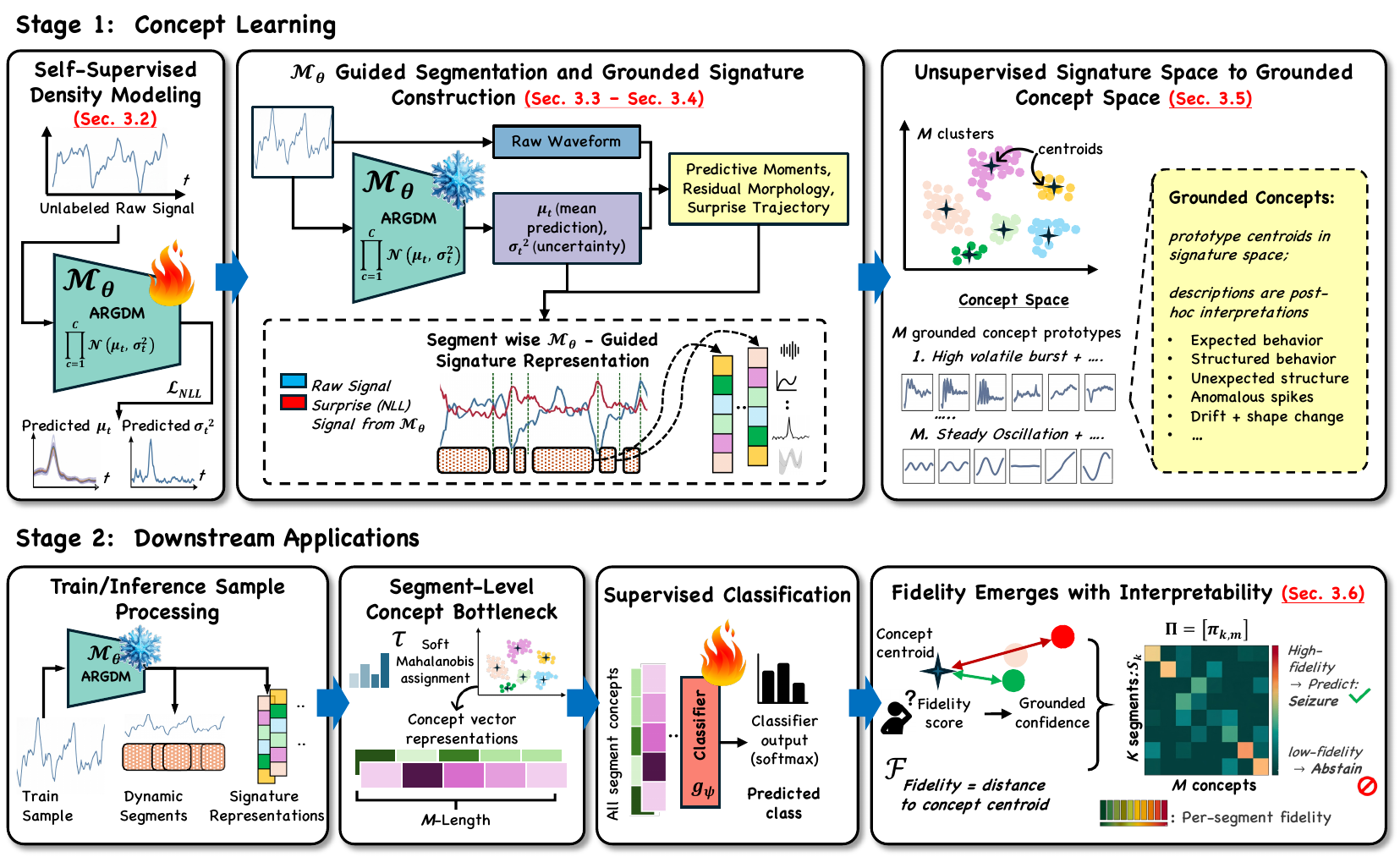}
    \caption{Overview of \texttt{ConceptTime}. 
    \textbf{Stage 1:} Autoregressive Gaussian Density Model (ARGDM) $\mathcal{M}_\theta$, trained once on raw signals, produces predictive surprise $s_t$ that drives length-constrained DP segmentation. Each segment is summarized by a signature vector that combines \emph{predictive moments}, \emph{residual morphology}, and the \emph{surprise trajectory}. Then, signatures are clustered into $M$ concept prototypes; textual labels are post hoc, not training signals.
    \textbf{Stage 2:} new inputs are segmented by $\mathcal{M}_\theta$, mapped to concepts by nearest-prototype assignment, and classified through the $K{\times}M$ concept tensor. Each segment's distance to its nearest prototype yields the fidelity score used for abstention.}
    \label{fig:method}
\end{figure}

\texttt{ConceptTime} factors classification into stages with disjoint training signals: a self-supervised density model produces every representation, clustering induces a non-parametric concept vocabulary, and a lightweight transformer head consumes labels. Each intermediate step is human-inspectable, and prototype distances provide a built-in fidelity score for selective prediction at no additional cost. Figure~\ref{fig:method} illustrates the 2-stage pipeline: (1) concept learning and (2) downstream applications.

\subsection{Problem Formulation}
A multivariate input is $x \in \mathbb{R}^{T \times C}$ with $T$ timesteps and $C$ channels. The classifier $f: \mathbb{R}^{T \times C} \to \{1, \ldots, Y\}$ is trained on $\{(x^{(i)}, y^{(i)})\}_{i=1}^N$. We fit the ARGDM and concept vocabulary on an unlabeled corpus $\mathcal{D}_{\text{u}}$; in our experiments, we use the training inputs without labels, but the framework supports any unlabeled corpus over the same signal space.

\subsection{Autoregressive Gaussian Density Model (ARGDM $\mathcal{M}_\theta$)}
\label{sec:method:density}

We train an autoregressive model $\mathcal{M}_\theta$ to predict the next-step diagonal Gaussian
\begin{equation}
    p_\theta(x_{t+1} \mid x_{\le t}) = \prod_{c=1}^{C} \mathcal{N}\!\bigl(x_{t+1, c};\, \mu_{t,c},\, \sigma_{t,c}^2\bigr),
    \qquad
    (\mu_t,\, \log \sigma_t^2) = \mathcal{M}_\theta(x_{\le t}),
\end{equation}
with a 4-layer causal Transformer. Training minimizes per-step Gaussian NLL augmented with a calibration regularizer $\bigl(\mathbb{E}[(x-\mu)^2/\sigma^2] - 1\bigr)^2$ that forces predicted variances to match realized residual energy. Calibration is not optional: without it, $\sigma_t^2$ collapses to a free fudge factor, and every downstream signature derived from it becomes uninterpretable. After this single self-supervised stage, we freeze $\mathcal{M}_\theta$ and reuse it for both segmentation and signature construction, without updating it during classification supervision (Stage 2).

\textbf{Per-timestep surprise.} The realised negative log-likelihood under $\mathcal{M}_\theta$,
\begin{equation}
    s_t = \frac{1}{C}\sum_{c=1}^{C} \tfrac{1}{2}\!\left(\log \sigma_{t,c}^2 + \frac{(x_{t,c} - \mu_{t,c})^2}{\sigma_{t,c}^2}\right),
\label{eq:surprise}
\end{equation}
drives segmentation. Unlike differential entropy $\tfrac{1}{2}\bigl(1 + \log 2\pi\sigma_t^2\bigr)$, which depends only on the predicted variance, $s_t$ additionally weights the realised residual $(x_t - \mu_t)^2$. It is therefore high precisely when the model was \emph{wrong despite being confident}, which is the regime transition a segmenter should mark.

\subsection{Length-Constrained Adaptive Segmentation}
\label{sec:method:patch}

Given $\{s_t\}_{t=1}^T$ and a target patch count $K$, we partition $x$ into $K$ non-overlapping segments by placing $K-1$ internal boundaries that maximize cumulative surprise subject to per-patch length constraints $[L_{\min}, L_{\max}]$. The dynamic program is
\begin{equation}
    B[k, t] = \max_{\substack{t':\; L_{\min} \le t - t' \le L_{\max}}}
    \bigl(B[k-1,\,t'] + s_t\bigr), \qquad B[0,0] = 0,
\label{eq:dp}
\end{equation}
with the same length constraint enforced on the trailing segment $T - t_{K-1}$. A monotone deque resolves the inner maximum in amortized constant time, giving $O(TK)$ overall. Length constraints prevent 2 failure modes of greedy threshold-based segmenters, such as EntroPE~\citep{abeywickrama2025entrope}: degenerate single-patch solutions on flat-surprise recordings and tail collapse when boundaries cluster at peaks.

\subsection{Predictive-Geometry Signatures}
\label{sec:method:signature}

Each patch $p_j = x_{t_j : t_{j+1}}$ is reduced to a signature $\phi_j \in \mathbb{R}^{2C+13}$ assembled from 3 families. All 3 are functions of $\mathcal{M}_\theta$'s output; no neural component sits between the signal and $\phi_j$.

\textbf{Predictive moments} ($2C+1$ dims): patch-mean of $\mu_t$ per channel, patch-mean of $\sigma_t^2$ per channel, and log patch length. These describe what the model expected, on average, and how confident it was.

\textbf{Residual morphology} (8 dims): FFT (Fast Fourier Transform) peak frequency~\citep{lubba2019catch22}, autocorrelations at lags 1–3, skewness, excess kurtosis, zero-crossing rate, and peak-to-peak range, all computed on the standardized residual, $r_t = (x_t - \mu_t) / \sigma_t$. Subtracting $\mu_t$ removes the predictable background dynamics that $\mathcal{M}_\theta$ already explains; dividing by $\sigma_t$ amplifies deviations in confident regions and damps them in uncertain ones. The resulting shape descriptors therefore reflect class-discriminative deviations from prediction rather than absolute amplitude or trend, which dominate the same statistics computed on raw $x_t$.

\textbf{Surprise trajectory} (4 dims): the within-patch position of $\arg\max_t s_t$ normalized to $[0, 1]$; the standard deviation of $\{s_t\}$; confidence-weighted surprise $\frac{1}{L}\sum_t s_t\bigl(1 - H_t/H_{\max}\bigr)$, which up-weights errors at instants where the model was most certain; and mean Mahalanobis residual energy $\frac{1}{L}\sum_t \|r_t\|^2$. Together, these specify \emph{when within the patch} and \emph{how severely} the model was surprised. Two patches with identical mean surprise but different temporal profiles (uniform versus a single late spike) produce distinct surprise-trajectory features and are routed to different concepts.

The 3 families capture mutually orthogonal views, namely, what was predicted (moments), what the residual looked like (morphology), and how surprise unfolded over time (trajectory). Signatures are standardized using 99th-percentile-clipped statistics estimated on $\mathcal{D}_{\text{u}}$, which prevents single-patch outliers from collapsing other dimensions to near-zero variance.

\subsection{Non-Parametric Concept Vocabulary}
\label{sec:method:concepts}

We fit a Gaussian Mixture Model (GMM) with diagonal covariance and $M$ components on the standardized signatures from $\mathcal{D}_{\text{u}}$, with $M \in \{8,16,32\}$ throughout. This range is small enough that the resulting concept-transition matrix remains human-readable, yet large enough to ensure class separability. Diagonal covariance handles the heterogeneous scales across the 3 signature families, and component-specific distance corrects for these scales at assignment time.

A patch with signature $\phi_j$ assigns to prototype $c_m$ via temperature-scaled soft assignment:
\begin{equation}
    \pi_{j,m} = \frac{\exp(-d(\phi_j, c_m)/\tau)}{\sum_{m'}\exp(-d(\phi_j, c_{m'})/\tau)},
    \qquad
    d(\phi_j, c_m) = \sqrt{(\phi_j - c_m)^\top \Sigma_m^{-1} (\phi_j - c_m)}.
\label{eq:assign}
\end{equation}
The concept tensor $\Pi \in \mathbb{R}^{K \times M}$ stacks $\{\pi_j\}_{j=1}^K$, and the nearest-prototype distance $d_j = \min_m d(\phi_j, c_m)$ is retained for the fidelity score below.

Because the entire path from signal to $\Pi$ is the composition of $\mathcal{M}_\theta$, the signature standardiser, and the GMM (none of which sees labels), the resulting concepts are immune to the concept-leakage failure mode in which a learned concept extractor smuggles label information into nominally interpretable activations~\citep{margeloiu2021concept}. Each prototype is a geometric centroid in signature space, with its nearest training patches serving as canonical exemplars.

\subsection{Classifier and Fidelity Score}
\label{sec:method:classifier}

A lightweight classifier $g_\psi$ maps the concept tensor $\Pi$ to class logits and is trained by cross-entropy. With a utilized range of $M$, $g_\psi$ has $\sim 10^3$–$10^4$ parameters; it is the only component trained under labeled supervision. Frozen $\mathcal{M}_\theta$, prototypes, and standardizer are shared across classifiers and tasks.

\textbf{Fidelity score.}
The same prototype distances $\{d_j\}_{j=1}^K$ already computed during concept assignment yield a built-in input-level fidelity measure:
\begin{equation}
    F(x) = \exp\!\left(-\frac{1}{K\tau} \sum_{j=1}^K d_j\right) \;\in\; (0, 1],
\label{eq:fidelity}
\end{equation}
where $\tau$ is set to the median patch-prototype distance on $\mathcal{D}_{\text{u}}$. A high $F(x)$ means every patch lies close to some concept centroid: the input is well described by the learned vocabulary. Low $F(x)$ means at least one segment falls in a region of signature space that no prototype covers; the input contains regimes the vocabulary cannot describe.

This is structurally different from softmax confidence. Softmax answers \emph{is the classifier confident in its prediction?}; $F(x)$ answers \emph{is the input describable in the concept space at all?}. The latter is a representation-level question that the classifier itself cannot pose, because by the time $g_\psi$ sees $\Pi$, all absolute distance information from $\phi_j$ to its nearest prototype has been normalized away by Equation~\ref{eq:assign}. Empirically (Section~\ref{sec:selective_results}), $F(x)$ achieves AUROC up to $0.99$ for OOD detection where softmax remains near chance.

\section{Experiments}
\label{sec:experiments}

We evaluate \texttt{ConceptTime} along 3 axes corresponding to the properties motivating the method, namely, \emph{accuracy}, \emph{interpretability}, and \emph{calibrated abstention}. Furthermore, we extend the analysis to cross-domain transfer-learning evaluation of the learned concept space.

\subsection{Experimental Setup}
\label{sec:setup_exp}

\textbf{Datasets.}
We evaluate on benchmarks spanning 4 domains, namely, (i) wearable sensing (UCI HAR~\citep{anguita2013public}), (ii) EEG-based biomedical monitoring (UCI Epilepsy~\citep{andrzejak2001indications}, merged to seizure/non-seizure), sleep staging (Sleep-EDF~\citep{goldberger2000physiobank,eldele2021tstcc}, Fpz-Cz channel), (iii) 10 UEA archive datasets~\citep{bagnall2018uea,wu2023timesnet}, and (iv) Fault Diagnosis (FD-A through FD-D~\citep{lessmeier2016condition,ragab2020adversarial}) for cross-domain transfer of the learned concepts.
Full dataset details are available in Appendix~\ref{app_data_detail}.

\textbf{Configuration.}
The frozen density model is a 4-layer causal transformer ($d=128$, $\lambda_{\mathrm{cal}}=0.25$, 50 epochs).
We use $(L_{\min},L_{\max})=(8,40)$, and $K$ and signature families are selected with validation split accuracies.
The classifier is a 2-layer transformer with $d_{\mathrm{cls}}=64$.
Results are the mean over 3 seeds (Appendix Table~\ref{tab:ci_tslanet_concepttime}); full hyperparameters in Appendix~\ref{app_hyper}.
We used a single NVIDIA RTX A5000 GPU for all experiments.

\textbf{Baselines.}
We compare against 4 categories.
\emph{Adaptive patching:} \texttt{ConceptTime}-Static (equal-length patches), PatchTST~\citep{nie2023a}, EntroPE~\citep{abeywickrama2025entrope}.
\emph{Black-box accuracy ceiling:} ROCKET~\citep{dempster2020rocket}, MiniRocket~\citep{dempster2021minirocket}, InceptionTime~\citep{ismail2020inceptiontime}, TS-TCC~\citep{eldele2021tstcc}, TimesNet~\citep{wu2023timesnet}, CrossFormer~\citep{zhang2023crossformer}, GPT4TS~\citep{zhou2023one}, TSLANet~\citep{eldele2024tslanet}; a single-layer MLP as a lower bound.
\emph{Interpretable methods:} catch22~\citep{lubba2019catch22}, ProSeNet~\citep{ming2019prosenet}, VQShape~\citep{wen2024abstracted}, TOTEM~\citep{talukder2024totem}.
\texttt{ConceptTime}-Dynamic combines adaptive patching with predictive-geometry concepts and is our primary method.

\subsection{Main Classification Results}
\label{sec:main_results}

\begin{table*}[t]
\centering
\caption{Classification accuracy (\%) under full, 1\%, and 5\% label budgets.
\capbest{Red}: best overall; \capsecond{Yellow}: second best overall;
\capint{Blue}: best interpretable method per column.
Shaded rows are ours.}
\label{tab:main_results}
\scriptsize
\setlength{\tabcolsep}{4.2pt}
\renewcommand{\arraystretch}{1.1}
\begin{tabular}{@{}ll ccc ccc ccc ccc@{}}
\toprule
& & \multicolumn{3}{c}{\textbf{UCI-HAR}}
  & \multicolumn{3}{c}{\textbf{Epilepsy}}
  & \multicolumn{3}{c}{\textbf{Sleep-EDF}}
  & \multicolumn{3}{c@{}}{\textbf{UEA (10 avg)}} \\
\cmidrule(lr){3-5} \cmidrule(lr){6-8} \cmidrule(lr){9-11} \cmidrule(l){12-14}
\textbf{Group} & \textbf{Method}
& 1\% & 5\% & Full
& 1\% & 5\% & Full
& 1\% & 5\% & Full
& 1\% & 5\% & Full \\
\midrule
\multirow{11}{*}{\rotatebox[origin=c]{90}{\textbf{Black-box}}}
& ROCKET & 82.21 & 89.65 & 93.91 & 93.68 & 92.25 & 97.15 & 70.21 & 73.67 & 76.38 & 55.69 & 59.87 & 72.30 \\
& MiniRocket & \bestall{86.55} & \bestall{95.63} & \bestall{97.31} & 94.71 & 96.86 & \secondall{98.26} & \bestall{79.53} & \bestall{82.15} & \bestall{85.91} & 51.90 & 54.89 & \secondall{73.80} \\
& InceptionTime & \secondall{86.38} & \secondall{91.03} & 95.32 & 95.14 & \secondall{97.28} & 98.21 & \secondall{77.64} & \secondall{81.11} & 84.62 & 48.37 & 50.80 & 67.42 \\
& TS-TCC & 70.23 & 75.82 & 93.72 & 91.20 & 93.20 & 97.12 & 75.26 & 78.98 & \secondall{85.67} & \bestall{62.37} & 63.87 & 72.71 \\
& TimesNet & 85.33 & 90.05 & 91.82 & 93.74 & 96.33 & 97.57 & 52.36 & 55.62 & 75.88 & \secondall{60.01} & 62.28 & 68.76 \\
& PatchTST & 28.27 & 73.46 & 85.71 & 92.57 & 96.67 & 93.65 & 65.19 & 67.72 & 69.93 & 49.43 & 50.90 & 70.52 \\
& GPT4TS & 84.58 & 87.36 & 89.37 & 82.71 & 83.93 & 85.51 & 69.36 & 71.22 & 76.22 & 59.31 & \secondall{65.47} & 68.31 \\
& CrossFormer & 69.89 & 74.12 & 77.35 & 71.55 & 75.28 & 76.18 & 51.28 & 55.68 & 53.98 & 59.63 & \bestall{67.54} & 70.58 \\
& TSLANet & 86.31 & 90.99 & \secondall{95.41} & \secondall{95.41} & 96.28 & 97.26 & 74.39 & 78.88 & 81.51 & 52.18 & 59.78 & \bestall{75.35} \\
& EntroPE & 71.29 & 74.14 & 84.41 & 85.62 & 87.61 & 89.94 & 70.95 & 71.84 & 75.12 & 56.37 & 58.27 & 61.06 \\
& MLP (1 layer) & 50.12 & 56.32 & 63.88 & 51.28 & 59.63 & 68.14 & 32.18 & 33.97 & 49.57 & 53.98 & 57.88 & 69.39 \\
\midrule
& TOTEM & 73.86 & 78.27 & 83.83 & 91.65 & 93.57 & 94.56 & 48.29 & 52.73 & 67.75 & 34.96 & 38.60 & 52.99 \\
& ProSeNet & 47.63 & 76.38 & 84.12 & 65.51 & 95.06 & 96.91 & 36.67 & 45.40 & 61.25 & 49.67 & 50.32 & 52.61 \\
& VQShape & 59.00 & 66.00 & 72.24 & 81.00 & 87.00 & 90.43 & 51.36 & 55.96 & 59.95 & 54.67 & 54.81 & 59.82 \\
& catch22 & 44.37 & 53.92 & 58.87 & 86.03 & 87.12 & 88.34 & 68.59 & 69.71 & 71.36 & 32.51 & 35.11 & 43.29 \\
\rowcolor{oursrow}
& \ct-Static & 69.81 & 75.29 & 83.27 & 94.24 & 94.13 & 95.47 & 68.26 & 69.92 & 73.36 & 55.98 & 51.29 & 67.82 \\
\rowcolor{oursrow}
\multirow{-6}{*}{\rotatebox[origin=c]{90}{\textbf{Interpretable}}}
& \ct-Dynamic & \bestint{74.60} & \bestint{79.60} & \bestint{88.91} & \bestall{97.91} & \bestall{98.01} & \bestall{98.39} & \bestint{70.21} & \bestint{71.92} & \bestint{78.15} & \bestint{58.50} & \bestint{60.10} & \bestint{71.82} \\
\bottomrule
\end{tabular}
\end{table*}

Across all 4 benchmarks and all label fractions, \texttt{ConceptTime}-Dynamic is the strongest interpretable method, with margins ranging from $1.5$ points (Epilepsy full-shot) to over $30$ points (HAR vs.\ catch22), while every other interpretable method degrades sharply at 1\% and 5\% labels.
At full supervision, \texttt{ConceptTime} matches or exceeds the strong black-box baselines on Epilepsy and trails by 2-8 points elsewhere, a controlled cost for enforcing a concept bottleneck.
Under label scarcity, this gap inverts: at 1\% labels, \texttt{ConceptTime} beats every black-box baseline on Epilepsy (97.91\% vs.\ TSLANet 95.41\%) and remains competitive on UEA datasets.
The classifier learns a mapping from concept activations to labels rather than from raw signals, making it intrinsically more sample-efficient (Table \ref{tab:main_results}). We further investigate more controlled, matched labeled and unlabeled training samples for the baselines in the few-label experiments, as discussed in Appendix~\ref{app:har_matched_seed2026}. 

\paragraph{Predictive geometry, not adaptive patching alone, drives the gain.}
Three ablations isolate the source of accuracy.
(1) Replacing dynamic patches with equal-length windows (\texttt{ConceptTime}-Static) loses 5.64 points on HAR and 2.92 on Epilepsy.
(2) Replacing DP segmentation with discrete categorical entropy (EntroPE) loses 4.5-8.5 points across benchmarks.
(3) Replacing predictive-geometry signatures with classical morphological descriptors on identical patches and clusters (catch22) reduces HAR by 30 points and UEA by 28 points.
The discriminative signal is the deviation from the prediction, not the raw waveform shape; adaptive patching is necessary but not sufficient.

\paragraph{Cross-domain transfer.}
We evaluate transfer across all 12 directed pairs of operating conditions in FD-A through FD-D (Table~\ref{tab:fd_transfer_maintext}). Panel A compares 3 \texttt{ConceptTime} settings. \emph{Direct} applies the source pipeline to the target domain without retraining. The 2 few-shot settings retain the source vocabulary and train a classifier with 5\% of the target labels. They differ in whether the density model is transferred from the source domain (Source-$\mathcal{M}_\theta$) or trained on unlabeled target inputs (Target-$\mathcal{M}_\theta$). Both use soft assignment to represent target examples in the source concept space. We compare against TSLANet with masked-reconstruction pretraining and full-model fine-tuning.

Direct transfer averaged 67.61\% accuracy across the 12 directions. Training only the target classifier raised the mean to 91.09\%. With the source vocabulary still fixed, using a target-trained density model raised accuracy by a further 3.28 percentage points to 94.37\%, the highest mean in Panel A. This setting ranked 1st on 7 directions. The source-density setting led on 2 directions (C$\to$A and D$\to$C), while TSLANet averaged 90.76\% and led on 3 (B$\to$D, D$\to$A, and D$\to$B).

Panel B examines whether adapting the vocabulary improves transfer when the source density model is frozen. All settings use the same 5\% labeled target subset. Averaged over 3 seeds, refitting the standardizer and GMM on unlabeled target inputs increased mean accuracy from 91.38\% to 91.74\% and improved 7 of the 12 directions. The effect varied by operating condition, with the largest gain on A$\to$D (+4.67 percentage points) and the largest decrease on B$\to$A ($-2.75$ points). Training both the density model and vocabulary on the target domain yielded 90.48\%.

\begin{table*}[t]
\centering
\caption{Target test accuracy (\%) across FD operating conditions. Within each panel,
\capbest{Red} and \capsecond{Yellow} indicate the highest and
second-highest accuracy for each direction, respectively.}
\label{tab:fd_transfer_maintext}
\scriptsize
\setlength{\tabcolsep}{2.5pt}
\renewcommand{\arraystretch}{1.1}
\begin{tabular}{@{}l *{12}{c} @{\hspace{5pt}} r r@{}}
\toprule
\textbf{Setting}
& \textbf{A$\to$B} & \textbf{A$\to$C} & \textbf{A$\to$D}
& \textbf{B$\to$A} & \textbf{B$\to$C} & \textbf{B$\to$D}
& \textbf{C$\to$A} & \textbf{C$\to$B} & \textbf{C$\to$D}
& \textbf{D$\to$A} & \textbf{D$\to$B} & \textbf{D$\to$C}
& \textbf{Mean} & \textbf{\#1st} \\
\midrule
\multicolumn{15}{@{}l}{\textbf{Panel A. Zero-shot and few-shot transfer}} \\
Direct, 0\%
& 36.25 & 44.46 & 29.58 & 50.62 & 81.82 & 98.57
& 52.05 & 87.84 & 94.20 & 57.19 & 93.60 & 85.15
& 67.61 & 0 \\
\rowcolor{oursrow}
Source-$\mathcal{M}_\theta$, 5\%
& \secondall{90.14} & 75.84 & \secondall{91.06}
& \secondall{90.40} & 84.02 & 98.53
& \bestall{96.66} & \secondall{97.43} & \secondall{97.62}
& \secondall{88.60} & 92.00 & \bestall{90.74}
& 91.09 & 2 \\
\rowcolor{oursrow}
Target-$\mathcal{M}_\theta$, 5\%
& \bestall{95.42} & \bestall{91.20} & \bestall{97.29}
& \bestall{93.73} & \bestall{90.54} & \secondall{98.72}
& \secondall{96.08} & \bestall{98.97} & \bestall{98.28}
& 87.94 & \secondall{95.06} & \secondall{89.17}
& 94.37 & 7 \\
TSLANet, 5\%
& 85.89 & \secondall{82.04} & 86.36
& 86.73 & \secondall{87.87} & \bestall{99.56}
& 89.11 & 96.30 & 96.00
& \bestall{90.65} & \bestall{99.85} & 88.78
& 90.76 & 3 \\
\midrule
\multicolumn{15}{@{}l}{\textbf{Panel B. Density model and vocabulary adaptation
with 5\% target labels}} \\
\rowcolor{oursrow}
Source-$\mathcal{M}_\theta$, source vocab.
& \secondall{94.00} & \secondall{87.70} & 92.45
& \bestall{89.53} & \secondall{87.49} & \secondall{94.27}
& \bestall{89.74} & \secondall{94.42} & \secondall{95.63}
& \secondall{88.60} & \bestall{92.00} & \secondall{90.74}
& 91.38 & 3 \\
\rowcolor{oursrow}
Source-$\mathcal{M}_\theta$, target vocab.
& \bestall{97.68} & \bestall{88.44} & \bestall{97.12}
& 86.78 & 86.14 & 92.60
& \secondall{88.54} & \bestall{95.96} & \bestall{96.22}
& \bestall{90.14} & 90.07 & \bestall{91.17}
& 91.74 & 7 \\
\rowcolor{oursrow}
Target-$\mathcal{M}_\theta$, target vocab.
& 90.86 & 87.66 & \secondall{95.06}
& \secondall{88.34} & \bestall{87.66} & \bestall{95.06}
& 88.34 & 90.86 & 95.06
& 88.34 & \secondall{90.86} & 87.66
& 90.48 & 2 \\
\bottomrule
\end{tabular}
\end{table*}

\subsection{Selective Prediction via Concept Fidelity}
\label{sec:selective_results}
The same prototype distances $\{d_j\}$ used for concept assignment yield a per-input fidelity score $F(x)$ (Eq.~\ref{eq:fidelity}) without additional computational cost.
We test whether this score is informative at the sample level (does it flag inputs the vocabulary cannot describe?) and at the patch level (does it rank the patches the classifier actually relies on?).

\begin{figure}[h!]
    \centering
    \includegraphics[width=0.75\linewidth]{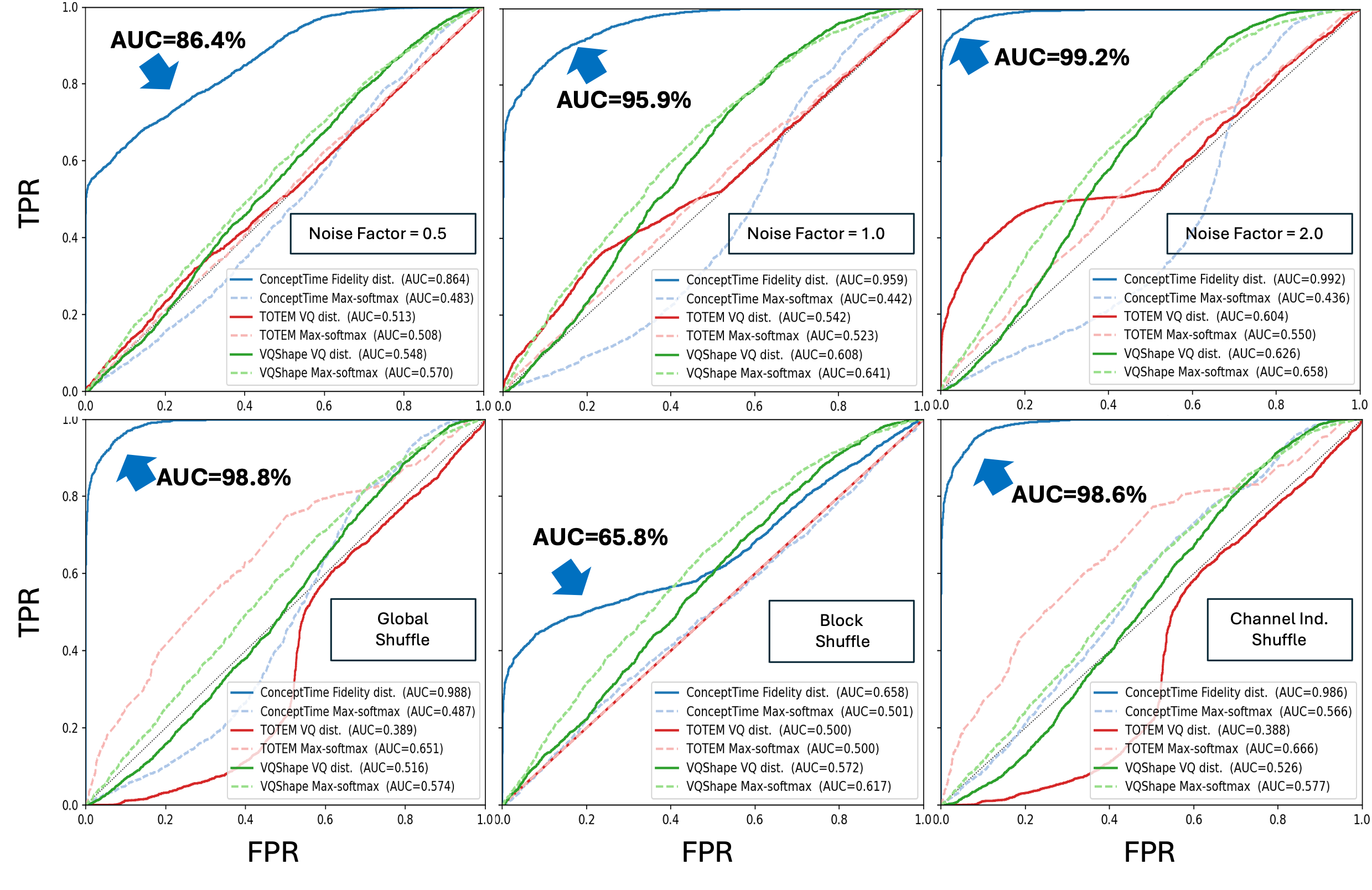}
    \caption{OOD detection on HAR under Gaussian-noise and temporal-shuffle perturbations.}
    \label{fig:ood}
\end{figure}

\paragraph{Sample-level OOD detection.} Figure~\ref{fig:ood} compares fidelity distance against max-softmax confidence on the same classifier and against prototype-distance scores from TOTEM and VQShape, under 3 Gaussian-noise intensities and 3 shuffle perturbations on HAR. Fidelity dominates across all 6 conditions, scaling smoothly with severity (AUROC $0.864 \!\to\! 0.992$ as noise factor grows from $0.5$ to $2.0$) while every other score remains near chance. The TOTEM and VQShape comparisons matter: prototype-based OOD detection is not a free property of codebook methods. It requires prototypes anchored in predictive-distribution geometry rather than reconstruction-based codes. We extend OOD detection beyond synthetic perturbations to a real-world FD dataset, detecting OOD under shifting operating conditions. We explain the details in Appendix~\ref{app:fd_natural_ood}. This OOD detection performance on FD further strengthens the capabilities of the by-design fidelity score of \texttt{ConceptTime}.

\begin{figure}[h!]
    \centering
    \includegraphics[width=0.8\linewidth]{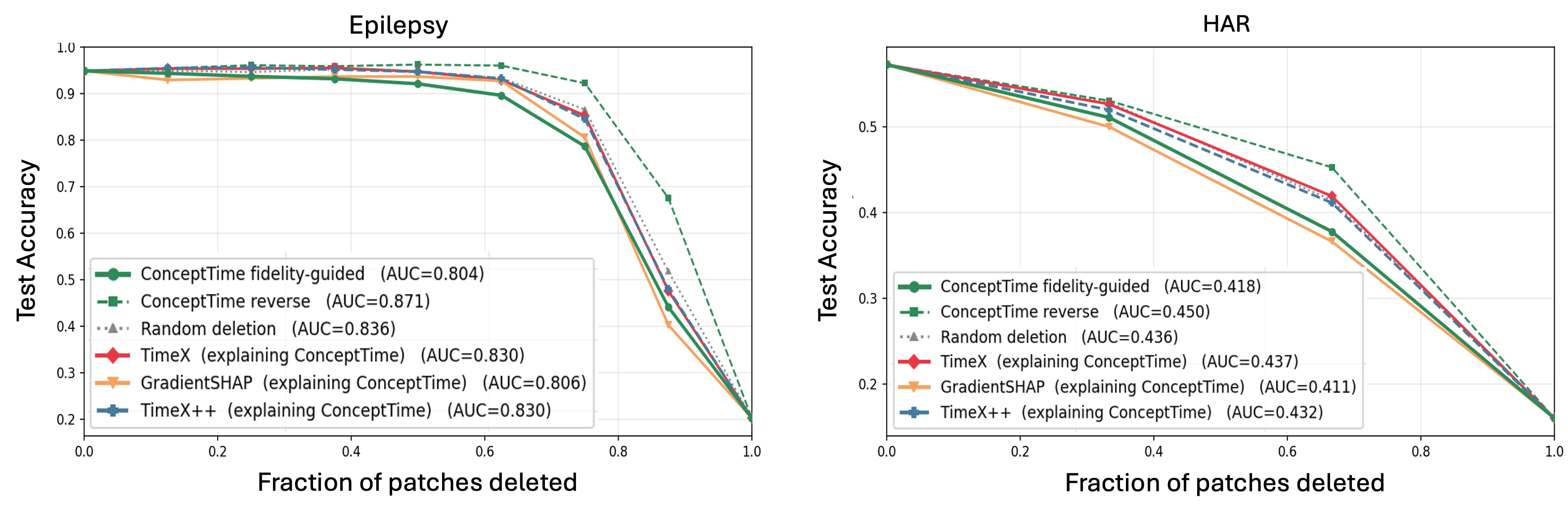}
    \caption{Deletion-based faithfulness on Epilepsy and HAR.
    Patches are removed in ranked order; lower AUC indicates more faithful attribution.
    }
    \label{fig:timex}
\end{figure}

\paragraph{Patch-level deletion faithfulness.} We rank patches by fidelity, progressively delete the highest-ranked ones, and measure accuracy decay (Algorithm~\ref{alg:deletion_faithfulness}); lower AUC means the deleted patches were more important.
We compare against 3 post hoc explainers (GradientSHAP~\citep{shap}, TimeX~\citep{queen2023encoding}, TimeX++~\citep{liu2024timex}) applied to the same \texttt{ConceptTime} classifier, as well as random and reverse-fidelity baselines. Figure~\ref{fig:timex} shows that fidelity-guided deletion achieves the lowest or near-lowest AUC on both Epilepsy and HAR, and the reverse baseline is substantially weaker, confirming the ranking is non-arbitrary. 
Mask-based methods (TimeX, TimeX++) cannot improve on random when explaining a concept-bottleneck classifier: reverse-engineering attributions through input perturbation is unreliable when the model's decision pathway is already structured.

Fidelity achieved the lowest deletion AUC on Epilepsy (0.804), while GradientSHAP led on HAR (0.411 versus 0.418 for fidelity). Both outperformed random deletion on both datasets.

\paragraph{Fidelity and classifier confidence.}
Max-softmax measures how strongly the classifier favors its predicted class given the patch-to-concept assignments. These assignments depend only on relative distances: adding the same constant to all prototype distances for a patch leaves its assignment unchanged (Eq.~\ref{eq:assign}). Fidelity uses the mean distance from each patch signature to its nearest prototype, so it can distinguish inputs with identical assignments but different distances to the learned concepts. It reuses distances computed during assignment and requires no additional training (Appendix~\ref{app:selective_epilepsy}).

\subsection{Ablation Study}
\label{sec:ablations}
Table~\ref{tab:ablations_all} reports full-supervision test accuracy when varying one component at a time. For each dataset, the signature ablation fixes $M=32$, and the vocabulary ablation fixes the residual-aware signature (Surp.f.). Within each sweep, all other settings are held fixed. Accuracies can therefore differ from Table~\ref{tab:main_results} when the main experiment uses a different configuration. In Table~\ref{tab:ablations_all}, Dist., Morph., and Mom.\ denote distributional features, raw-signal morphology, and predictive moments, respectively. Full combines all 4 feature families ($2C+17$ dimensions), while Surp.f.\ combines 3 residual-aware feature families ($2C+13$ dimensions). Ent.\ denotes the entropy-only signature in the signature ablation and the entropy-based boundary signal in the boundary ablation. KL denotes the KL divergence between consecutive predictive Gaussians, and Surp.\ denotes per-timestep surprise.

\begin{table*}[h!]
\centering
\caption{Unified ablation on UCI-HAR, Epilepsy, and Sleep-EDF (full-supervision accuracy, \%).
Within each ablation group, \capbest{Red} marks the best value per dataset and \capsecond{Yellow} the second best.}
\label{tab:ablations_all}
\scriptsize
\begin{tabular}{@{}l cccccc @{\hspace{8pt}} ccccc @{\hspace{8pt}} ccc@{}}
\toprule
& \multicolumn{6}{c}{\textbf{Signature components}}
& \multicolumn{5}{c}{\textbf{Vocabulary size ($M$)}}
& \multicolumn{3}{c@{}}{\textbf{Boundary signal}} \\
\cmidrule(lr){2-7} \cmidrule(lr){8-12} \cmidrule(l){13-15}
\textbf{Dataset}
& Dist. & Ent. & Morph. & Mom. & Full & Surp.f.
& 8 & 16 & 32 & 64 & 128
& Ent. & KL & Surp. \\
\midrule
UCI-HAR
& 72.8 & 61.2 & 42.7 & \bestall{87.7} & 82.8 & \secondall{85.7}
& 80.2 & 84.4 & 85.7 & \secondall{86.3} & \bestall{86.8}
& \bestall{83.4} & 81.6 & \secondall{82.5} \\
Epilepsy
& \secondall{97.2} & 96.1 & 94.6 & 96.4 & 96.7 & \bestall{98.0}
& 98.1 & 98.2 & 98.0 & \secondall{98.4} & \bestall{98.5}
& 97.4 & \secondall{97.6} & \bestall{98.1} \\
Sleep-EDF
& 72.3 & 65.4 & 59.0 & 74.4 & \bestall{78.3} & \secondall{74.8}
& 73.7 & 75.0 & 74.0 & \secondall{75.4} & \bestall{76.2}
& 73.8 & \bestall{75.0} & \secondall{74.5} \\
\bottomrule
\end{tabular}
\end{table*}

\paragraph{Signature components.}
The most informative patch features vary across datasets. On UCI-HAR, predictive moments achieve 87.7\% accuracy, 45.0 percentage points above raw-signal morphology on the same adaptive patches. On Epilepsy, distributional features are the strongest individual family (97.2\%), while the residual-aware signature achieves the highest overall accuracy (98.0\%). On Sleep-EDF, combining all 4 families gives the best result (78.3\%).

\paragraph{Vocabulary size.}
The largest tested vocabulary ($M=128$) achieves the highest accuracy on all 3 datasets. The benefit of a larger vocabulary is most pronounced on UCI-HAR. At $M=32$, however, accuracy is within 2.2 percentage points of the best result on each dataset. We use $M=32$ as the default to keep the concept vocabulary small enough to inspect.

\paragraph{Boundary signal.}
Accuracy varies by at most 1.8 percentage points across boundary signals within each dataset. Entropy performs best on UCI-HAR, surprise on Epilepsy, and KL-shift on Sleep-EDF. We use surprise as the default because it places boundaries using the same predictive deviations described by the residual-aware signature.

\subsection{Qualitative Analysis}
\label{sec:concept_quality}

\begin{figure*}[h!]
    \centering
    \includegraphics[width=\linewidth]{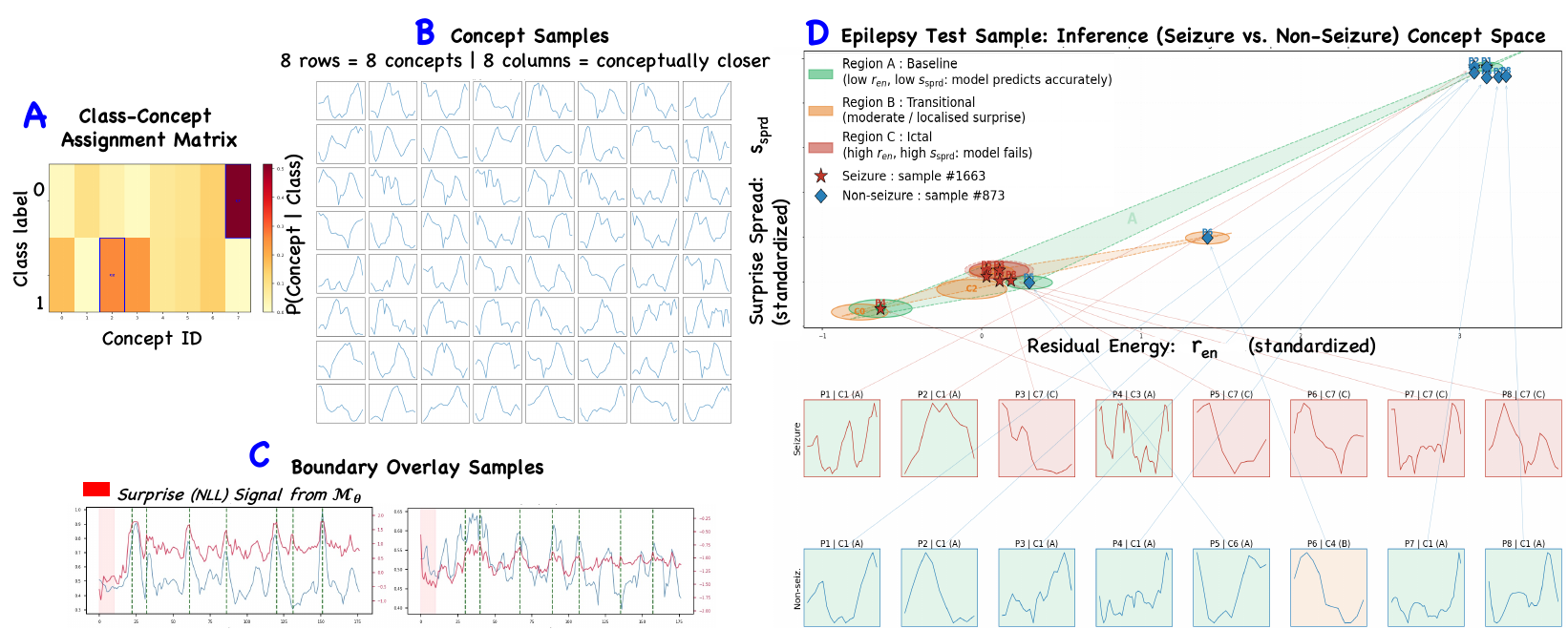}
    \caption{Learned concepts, patch boundaries, and predictive signatures on Epilepsy.}
    \label{fig:qualitative}
\end{figure*}
 
Figure~\ref{fig:qualitative} shows how the learned concepts relate to seizure activity. \bestint{A:} Although the GMM is fitted without labels, $P(\mathrm{concept}\mid\mathrm{class})$ peaks at C7 for non-seizure and C2 for seizure. \bestint{B:} Each row shows the 8 patches closest to one prototype (C0-C7), revealing recurring waveform patterns within concepts. \bestint{C:} Several DP boundaries coincide with peaks in predictive surprise $s_t$ (red). \bestint{D:} Residual energy $r_{en}$ and surprise spread $s_{sprd}$ show where the 2 examples overlap and differ. 5 of 8 seizure patches ($\star$) fall in Region C, where the density model predicts poorly, while 7 of 8 non-seizure patches ($\diamond$) fall in Region A, where it predicts accurately. Both examples contain Region A patches with different waveforms, illustrating how predictive signatures can group visually distinct segments (Figure~\ref{fig:teaser}). Further examples appear in Appendix~\ref{app:vizz}.

\section{Conclusion}
We introduced \texttt{ConceptTime}, a concept-bottleneck framework that resolves the segmentation-concept circularity for time series through a single self-supervised primitive, \textit{predictive surprise}.
The same frozen model sets adaptive segment boundaries, grounds a non-parametric concept vocabulary, and provides prototype distances that also serve as a built-in fidelity score.
Across diverse classification and cross-domain transfer benchmarks, \texttt{ConceptTime} maintains competitive accuracy with black-box methods, dominates all interpretable baselines, including in few-shot settings. The by-design fidelity score remains competitive with softmax, prototype-distance, and post-hoc explainers on OOD detection and deletion faithfulness.
Interpretability, label efficiency, and reliable abstention emerge jointly when concept grounding is anchored in predictive surprise rather than human annotation.

\paragraph{Limitations and Future Work.}
We fix the vocabulary size in advance, and the meaning of each learned prototype depends on human interpretation after training. Hierarchical vocabularies could capture patterns at multiple levels of detail, while online updates could help the vocabulary adapt as signal patterns change. Low-fidelity segments could also guide active relabeling by identifying examples for human review. Beyond classification, we plan to explore whether these concepts can help discover operating regimes, locate anomalies, and support decisions that users can inspect.

\section*{Acknowledgments}
We thank the anonymous reviewers for their helpful feedback. Sachith Abeywickrama acknowledges support from the Agency for Science, Technology and Research (A*STAR), Singapore, through the Singapore International Graduate Award (SINGA). This research is supported by the Ministry of Education, Singapore, under the MOE Tier 1 grant (SKI 2021\_08\_03), and by the Kwan Im Thong Hood Cho Temple Chair Professorship. This research is also supported by the KU Research Center for Advanced Intelligent Systems (KU-AIS) at Khalifa University of Science and Technology.

\bibliographystyle{plainnat}
\bibliography{neurips_2026}

\newpage
\appendix
\section*{Appendix}
\section{Extended Related Work}
\label{app_extended_realted_work}

\subsection{Concept bottleneck models}

Concept bottleneck models (CBMs)~\citep{koh2020concept} route predictions through human-interpretable intermediate concepts, enabling inspection and intervention.
Classical CBMs require dense concept annotations, motivating a line of work on \emph{label-free} variants that source concepts from large pretrained models.
Label-free CBM~\citep{oikarinen2023labelfree} and Language-in-a-Bottle~\citep{yang2023language} use CLIP and LLMs to mine concept names; VLG-CBM~\citep{srivastava2024vlg} grounds concepts via open-vocabulary detectors; recent work eliminates the CLIP dependency through unsupervised classifier-distribution alignment~\citep{sammani2025clipfree}.
Other extensions improve concept robustness~\citep{margeloiu2021concept, sawada2022concept}, support counterfactual reasoning~\citep{dominici2025counterfactual}, and add concept locality~\citep{jeon2025locality}.

These approaches motivate learning a vocabulary directly from time-series recordings. In \texttt{ConceptTime}, a frozen density model supplies the features used to fit prototypes without concept annotations. Direct assignment to those prototypes avoids a separately trained concept predictor, one possible source of concept leakage~\citep{margeloiu2021concept}.

\paragraph{Time-series concepts.}

\citet{wu2022learning} use predefined statistical summaries as concepts for clinical signals; \citet{sprang2025enforcing} apply CKA-aligned concepts to forecasting transformers; \citet{stelle2025} use Signal Temporal Logic formulae as interpretable predicates.

These methods begin with a specified vocabulary. In time series, the duration and identity of a pattern can both be unknown. \texttt{ConceptTime} uses the same forecaster to locate segment boundaries and characterize the resulting segments.

\subsection{Segmentation and representation learning}

\paragraph{Adaptive patching.}

Patch-based transformers such as PatchTST have demonstrated the value of segmenting time series into patches.
The Byte Latent Transformer (BLT)~\citep{pagnoni2025blt} extends this idea by placing boundaries at points of high next-token entropy in language modeling.
EntroPE~\citep{abeywickrama2025entrope} adapts the idea to time series, training a discrete autoregressive model and segmenting via a greedy dual-threshold rule on categorical entropy from quantized tokens.

ReinPatch~\citep{wu2026dynamic} learns patch boundaries through reinforcement learning, and TimeSqueeze~\citep{ankireddy2026timesqueeze} uses content-aware compression for forecasting. In \texttt{ConceptTime}, a continuous Gaussian forecaster provides predictive means, variances, and realized surprise. By default, a length-constrained dynamic program selects a fixed number of patches using surprise. The same forecaster supplies the statistics used to describe each patch.

\paragraph{Discrete tokenization.}

TOTEM~\citep{talukder2024totem} learns a VQ-VAE codebook for time-series patches; VQShape~\citep{wen2024abstracted} learns a shape-level codebook with subsequence reconstruction; Chronos~\citep{ansari2024chronos} tokenizes scaled values into a 4096-symbol vocabulary for autoregressive forecasting.

\texttt{ConceptTime} clusters signatures whose coordinates summarize expected behavior, residual morphology, and surprise within a patch. These features and nearby training patches give users two ways to inspect a prototype.

\paragraph{Predictive representations.}

Our use of a probabilistic backbone for representation learning connects to predictive state representations, regime discovery, and information-bottleneck approaches for sequential data~\citep{luo2023time, choi2023conditional, ullmann2023multivariate}.
Time-series foundation models such as MOMENT~\citep{goswami2024moment} and Chronos~\citep{ansari2024chronos} demonstrate that self-supervised pretraining on raw signals yields useful general-purpose representations.

In \texttt{ConceptTime}, predictive statistics connect self-supervised pretraining to the downstream classifier by defining both segments and the vocabulary used to encode them.

\subsection{Prototypes, explanations, and selective prediction}

\paragraph{Prototype and shapelet classifiers.}

A long line of work uses prototypes or shapelets as interpretable units for time-series classification~\citep{ye2009time, ming2019prosenet, gee2019explaining, xie2024prototype}.
ProSeNet~\citep{ming2019prosenet} learns a small library of prototype sequences with case-based reasoning.
Shapelet methods~\citep{ye2009time, grabocka2014learning} discover discriminative subsequences and classify by distance.
ECATS~\citep{ferfoglia2024ecats} and related work~\citep{wen2025shedding, wen2024abstracted} use symbolic predicates and abstracted shapes.

UniShape~\citep{liu2026unishape} uses supervised pretraining with prototypes of multi-scale windowed shapes for transferable classification. \texttt{ConceptTime} retains the local, example-based interpretation of prototype methods and adds the preceding temporal context through the forecaster. Two patches with similar waveforms can therefore receive different concept assignments.

\paragraph{Post-hoc explanations.}

A complementary line of work explains pretrained black-box classifiers post hoc through saliency, attribution, or perturbation~\citep{queen2023encoding, liu2024timex, crabbe2021explaining, huang2026shapex}.
TimeX~\citep{queen2023encoding} and TimeX++~\citep{liu2024timex} apply information bottleneck principles to extract sub-instance explanations; ShapeX~\citep{huang2026shapex} aligns Shapley attribution with shapelet segments.

These methods identify evidence used by a pretrained classifier. In \texttt{ConceptTime}, the classifier receives the sequence of concept assignments as its input.

\paragraph{Selective prediction.}

Selective prediction has been studied through softmax confidence~\citep{geifman2019selective}, ensemble disagreement~\citep{lakshminarayanan2017simple}, and distance-based OOD scoring~\citep{lee2018simple, sun2022out}; recent work explores prototype-based OOD detection with mixtures~\citep{lu2024learning}, and distance to in-distribution centroids.
In vision, \emph{concept-based} selective prediction has emerged through Conceptual Safeguards~\citep{joren2024classification} and Sidecar-CBM~\citep{lockhart2022learn}, which propagate concept-level uncertainty into the abstention decision, and through stochastic concept representations~\citep{vandenhirtz2024stochastic, kim2023probcbm} and selective concept models~\citep{schrodi2025selective}.

Time-series research also studies generalization to out-of-distribution inputs~\citep{lu2024diversify}. Prototype distance in \texttt{ConceptTime} measures how closely segment signatures match the learned vocabulary. This supplies a score for abstention without fitting a separate confidence model.

\paragraph{Comparison of approaches.}

Table~\ref{tab:capability_comparison} summarizes the methods discussed above. \texttt{ConceptTime} brings together adaptive segmentation, concept discovery without concept annotations, and a prototype-distance score for selective prediction.

\begin{table}[!htbp]
\centering
\caption{
Comparison of major directions in interpretable time-series classification.
\cmark\ = direct support; \xmark\ = absent; \pmark\ = partial or with caveats.
}
\label{tab:capability_comparison}
\tiny
\setlength{\tabcolsep}{3.5pt}
\renewcommand{\arraystretch}{1.1}
\begin{tabularx}{\textwidth}{>{\raggedright\arraybackslash}X 
  >{\centering\arraybackslash}p{1.2cm}
  >{\centering\arraybackslash}p{1.2cm}
  >{\centering\arraybackslash}p{1.3cm}
  >{\centering\arraybackslash}p{1.4cm}
  >{\centering\arraybackslash}p{1.5cm}
  >{\centering\arraybackslash}p{1.3cm}}
\toprule
\textbf{Property} &
\textbf{Post-hoc} &
\textbf{CBMs} &
\shortstack{\textbf{Adaptive}\\\textbf{Patch}} &
\shortstack{\textbf{Prototype}\\\textbf{/ Shapelet}} &
\shortstack{\textbf{IB / Repr.}\\\textbf{Learning}} &
\shortstack{\textsc{\textbf{ConceptTime}}\\\textbf{(Ours)}} \\
\midrule
Decision routes through concepts   & \xmark & \cmark & \xmark & \pmark & \xmark & \cmark \\
No external semantic supervision required                       & \cmark & \pmark & \cmark & \cmark & \cmark & \cmark \\
Concepts discovered from data, not predefined                   & \pmark & \pmark & --     & \pmark & \cmark & \cmark \\
Adaptive segmentation                                            & \xmark & \xmark & \cmark & \xmark & \xmark & \cmark \\
Local, segment-level explanations                                & \pmark & \pmark & \xmark & \cmark & \xmark & \cmark \\
Built-in abstention without auxiliary head                       & \xmark & \pmark & \xmark & \xmark & \pmark & \cmark \\
Competitive supervision accuracy                                 & \cmark & \pmark & \cmark & \pmark & \cmark & \cmark \\
Strong few-shot performance                                      & \xmark & \xmark & \xmark & \xmark & \pmark & \cmark \\
\bottomrule
\end{tabularx}
\begin{flushleft}
\scriptsize
\textit{Notes.}
Post-hoc: TimeX, TimeX++, ShapeX~\citep{queen2023encoding,liu2024timex,huang2026shapex}.
CBMs: vision-language CBMs and time-series CBMs~\citep{koh2020concept,oikarinen2023labelfree,wu2022learning,sprang2025enforcing}; 
Adaptive Patch: BLT, EntroPE~\citep{pagnoni2025blt,abeywickrama2025entrope}; the dash indicates the property is outside scope (these methods are not designed to expose concepts).
Prototype/Shapelet: ProSeNet, ECATS, shapelets~\citep{ming2019prosenet,ferfoglia2024ecats,ye2009time}.
IB/Repr.: information-bottleneck and contrastive representation methods~\citep{luo2023time,choi2023conditional}.
\end{flushleft}
\end{table}

\FloatBarrier

\section{Datasets and Training Setup}
\label{app:datasets_training}

\subsection{Benchmarks and data splits}
\label{app_data_detail}

Table~\ref{tab:datasets_appendix} summarizes the main benchmark groups used in our experiments.

\begin{itemize}
    \item \textbf{UCI HAR.}
    The UCI Human Activity Recognition dataset contains smartphone inertial recordings from 30 subjects performing 6 daily activities: walking, walking upstairs, walking downstairs, sitting, standing, and lying down.
    The original study used a waist-mounted Samsung Galaxy S2 to record three-axis acceleration and angular velocity.
    Each sample is represented as a 9-channel window of length $T=128$, and the task is 6-way activity classification.

    \item \textbf{UCI Epilepsy.}
    The Epileptic Seizure Recognition dataset consists of single-channel EEG recordings.
    The original recordings span 23.6 seconds each.
    The original dataset contains 5 classes, but only one class corresponds to epileptic seizure activity.
    Following common practice, we merge the remaining 4 non-seizure classes into a single class and formulate the task as binary seizure detection.
    Each classification example contains $T=178$ time steps from an original recording.

    \item \textbf{Sleep-EDF.}
    Sleep-EDF contains whole-night polysomnography recordings from the PhysioBank database.
    The task is sleep-stage classification into 5 classes: Wake, N1, N2, N3, and REM.
    Following prior work, we use the single EEG channel Fpz-Cz sampled at 100Hz.
    Each input window has length $T=3000$, corresponding to a 30-second EEG epoch.

    \item \textbf{UEA multivariate archive subset.}
    We evaluate on a subset of 10 datasets from the UEA multivariate time-series classification archive.
    These datasets cover diverse domains, including alcohol industry measurements, face detection, handwriting, heartbeat monitoring, speech, transportation, EEG self-regulation, spoken digit recognition, and gesture recognition.
    They vary substantially in dimensionality, sequence length, and number of classes, making them useful for testing whether the learned concept representation generalizes beyond a single sensing modality.

    \item \textbf{Fault Diagnosis.}
    The Fault Diagnosis dataset is a real-world bearing fault classification benchmark collected under 4 working conditions, denoted FD-A, FD-B, FD-C, and FD-D.
    We treat each working condition as a separate domain because signal characteristics differ across operating regimes.
    All domains share the same label space (3 classes): healthy, inner fault, and outer fault.
    We use this dataset for cross-domain transfer experiments, where a model is trained on one operating condition and adapted or evaluated on another.
\end{itemize}

\begin{table*}[!htbp]
\centering
\caption{Summary of datasets used in the experiments.}
\label{tab:datasets_appendix}
\footnotesize
\setlength{\tabcolsep}{4pt}
\renewcommand{\arraystretch}{1.12}
\begin{tabular}{l l c c c c l}
\toprule
\textbf{Dataset} 
& \textbf{Domain / Modality} 
& \textbf{\#Train} 
& \textbf{\#Test} 
& \textbf{\#Ch.} 
& \textbf{$T$} 
& \textbf{Task / Classes} \\
\midrule
UCI HAR~\citep{anguita2013public} 
& Wearable inertial sensing 
& 7352 
& 2947 
& 9 
& 128 
& 6 activities \\

UCI Epilepsy~\citep{andrzejak2001indications} 
& EEG 
& 9200 
& 2300 
& 1 
& 178 
& Seizure / non-seizure \\

Sleep-EDF~\citep{goldberger2000physiobank,eldele2021tstcc} 
& Sleep EEG 
& 25612 
& 8910 
& 1 
& 3000 
& 5 sleep stages \\

UEA subset~\citep{bagnall2018uea,wu2023timesnet} 
& Mixed multivariate time series 
& Varies 
& Varies 
& Varies 
& Varies 
& 10 classification datasets \\

Fault Diagnosis~\citep{lessmeier2016condition,ragab2020adversarial} 
& Bearing fault diagnosis 
& 8184 
& 2728 
& 1 
& 5120 
& 3 classes; 4 transfer domains \\
\bottomrule
\end{tabular}
\end{table*}

\subsection{UEA archive subset}

Table~\ref{tab:uea_subset_appendix} lists the ten UEA datasets. We use their official training and test splits and, when needed, reserve validation examples from the training split.

\begin{table*}[!htbp]
\centering
\caption{
UEA multivariate archive subset used in our evaluation. 
The split column follows the format \#train/\#val/\#test before any validation split is carved from the training set.
}
\label{tab:uea_subset_appendix}
\footnotesize
\setlength{\tabcolsep}{4pt}
\renewcommand{\arraystretch}{1.12}
\begin{tabular}{l c c c l}
\toprule
\textbf{Dataset} 
& \textbf{\#Ch.} 
& \textbf{$T$} 
& \textbf{Split} 
& \textbf{Domain} \\
\midrule
EthanolConcentration 
& 3 
& 1751 
& 261 / 0 / 263 
& Alcohol industry \\

FaceDetection 
& 144 
& 62 
& 5890 / 0 / 3524 
& Face / sensor signals \\

Handwriting 
& 3 
& 152 
& 150 / 0 / 850 
& Handwriting \\

Heartbeat 
& 61 
& 405 
& 204 / 0 / 205 
& Heartbeat \\

JapaneseVowels 
& 12 
& 29 
& 270 / 0 / 370 
& Voice \\

PEMS-SF 
& 963 
& 144 
& 267 / 0 / 173 
& Transportation \\

SelfRegulationSCP1 
& 6 
& 896 
& 268 / 0 / 293 
& Health / EEG \\

SelfRegulationSCP2 
& 7 
& 1152 
& 200 / 0 / 180 
& Health / EEG \\

SpokenArabicDigits 
& 13 
& 93 
& 6599 / 0 / 2199 
& Voice \\

UWaveGestureLibrary 
& 3 
& 315 
& 120 / 0 / 320 
& Gesture recognition \\
\bottomrule
\end{tabular}
\end{table*}

\subsection{Training hyperparameters}
\label{app_hyper}

We used the same Transformer classifier architecture across datasets. We trained the classifier for 200 epochs with a batch size of 64 using AdamW with an initial learning rate of $3\times10^{-3}$, weight decay of $0.01$, and a cosine annealing schedule.

For \texttt{ConceptTime}-Dynamic, we considered vocabulary sizes $M\in\{8,16,32\}$, surprise- and entropy-based patch boundaries, and signature variants including the residual-aware signature and predictive moments alone. Here $M$ denotes the number of concepts in the vocabulary. The vocabulary ablation additionally evaluated $M=64$ and $M=128$, and the boundary ablation included KL-shift (Table~\ref{tab:ablations_all}). \texttt{ConceptTime}-Static used fixed patch boundaries.

\FloatBarrier

\section{Classification Accuracy and Label Efficiency}
\label{app:classification_results}

\subsection{Accuracy across training seeds}
\label{app:seed_accuracy}

Table~\ref{tab:ci_tslanet_concepttime} reports mean test accuracy and standard deviation across three random seeds for \texttt{ConceptTime}-Dynamic and TSLANet, the reference black-box classifier. \texttt{ConceptTime} has higher mean accuracy on Epilepsy at all three label budgets. On UEA, its mean is higher at 1\% and 5\% labels, while TSLANet leads with full supervision. TSLANet leads at all three budgets on UCI-HAR and Sleep-EDF.

\begin{table}[!htbp]
\centering
\caption{
Test accuracy (\%, mean $\pm$ standard deviation across three seeds) for TSLANet and \texttt{ConceptTime}-Dynamic.
}
\label{tab:ci_tslanet_concepttime}
\scriptsize
\setlength{\tabcolsep}{4pt}
\renewcommand{\arraystretch}{1.08}
\begin{tabular}{llcc}
\toprule
\textbf{Dataset} & \textbf{Labels} & \textbf{TSLANet} & \textbf{\texttt{ConceptTime}-Dynamic} \\
\midrule
\multirow{3}{*}{UCI-HAR}
& Full & 95.41 $\pm$ 0.63 & 88.91 $\pm$ 3.40 \\
& 1\%  & 86.31 $\pm$ 1.42 & 74.60 $\pm$ 1.76 \\
& 5\%  & 90.99 $\pm$ 1.08 & 79.60 $\pm$ 1.31 \\
\midrule
\multirow{3}{*}{Epilepsy}
& Full & 97.26 $\pm$ 0.51 & \underline{98.39 $\pm$ 0.35} \\
& 1\%  & 95.41 $\pm$ 1.19 & \underline{97.91 $\pm$ 0.96} \\
& 5\%  & 96.28 $\pm$ 0.87 & \underline{98.01 $\pm$ 0.71} \\
\midrule
\multirow{3}{*}{Sleep-EDF}
& Full & 81.51 $\pm$ 1.24 & 78.15 $\pm$ 1.53 \\
& 1\%  & 74.39 $\pm$ 2.31 & 70.21 $\pm$ 1.84 \\
& 5\%  & 78.88 $\pm$ 1.39 & 71.92 $\pm$ 1.67 \\
\midrule
\multirow{3}{*}{UEA avg.}
& Full & 75.35 $\pm$ 0.82 & 71.82 $\pm$ 1.13 \\
& 1\%  & 52.18 $\pm$ 2.06 & 58.50 $\pm$ 2.24 \\
& 5\%  & 59.78 $\pm$ 1.58 & 60.10 $\pm$ 2.11 \\
\bottomrule
\end{tabular}
\end{table}

\subsection{Scaling with unlabeled pretraining data}
\label{app:unlabeled_scaling}

On Epilepsy, we vary the fraction of training inputs used to fit the density model, signature standardizer, and GMM. Within each seed, the 6 subsets are nested. At each label budget, the classifier uses the same labeled examples throughout the sweep. We use held-out validation data for selection and reserve the test set for evaluation.

\begin{table}[!htbp]
\centering
\caption{Unlabeled-data scaling on Epilepsy. Classification entries give mean test accuracy (\%) $\pm$ standard deviation across 3 seeds. The last column gives the density model's validation negative log-likelihood (NLL). In each accuracy column, \capbest{red} marks the highest mean and \capsecond{yellow} the second highest.}
\label{tab:app_unlabeled_scaling}
\small
\setlength{\tabcolsep}{4pt}
\begin{tabular}{@{}rccc@{}}
\toprule
Unlabeled fraction & 1\% labels & 5\% labels & Val.\ NLL $\downarrow$ \\
\midrule
1\%   & 95.64$\pm$ 0.16 & 96.74$\pm$0.50 & -1.422 \\
5\%   & 95.78$\pm$ 1.81 & 97.12$\pm$0.78 & -1.631 \\
10\%  & 96.46$\pm$ 0.93 & \bestall{97.42$\pm$0.37} & -1.839 \\
25\%  & 97.07$\pm$ 0.60 & 97.22$\pm$0.29 & -1.970 \\
50\%  & \secondall{97.33$\pm$ 0.41} & 97.19$\pm$0.25 & -2.223 \\
100\% & \bestall{97.36$\pm$ 0.35} & \secondall{97.30$\pm$0.26} & -2.233 \\
\bottomrule
\end{tabular}
\end{table}

Validation NLL decreases at every step (Table~\ref{tab:app_unlabeled_scaling}). With 1\% labels, increasing the unlabeled fraction from 1\% to 100\% raises accuracy by 1.72 percentage points, with only 0.03 points gained between 50\% and 100\%. With 5\% labels, the endpoint gain is 0.56 points, and the highest mean accuracy occurs at 10\% unlabeled data. By that fraction, the fitted GMM no longer contains near-empty components. Fidelity AURC varies non-monotonically across the sweep. We restrict the scaling comparison to Epilepsy because the preliminary UCI-HAR sweep covers only one seed.


\subsection{Effect of density model size}
\label{app:density_model_size}

We vary the depth and hidden dimension $d$ of the density model and measure downstream classification accuracy on Epilepsy and UCI-HAR
(Table~\ref{tab:app_density_model_size}). Parameter counts refer to the density model.

\begin{table}[!htbp]
\centering
\caption{Density model size and downstream test accuracy (mean $\pm$ standard deviation across 3 seeds). \capbest{red} marks the best value and \capsecond{yellow} the second best.}
\label{tab:app_density_model_size}
\small
\setlength{\tabcolsep}{4pt}
\begin{tabular}{@{}llrrrc@{}}
\toprule
Dataset & Size & Layers & $d$ & Parameters & Accuracy (\%) \\
\midrule
Epilepsy & Tiny    & 2 & 64  & 110,149   & \bestall{98.65$\pm$0.47} \\
         & Default & 4 & 128 & 810,629   & \secondall{98.39$\pm$0.35} \\
         & Larger  & 6 & 128 & 1,204,357 & 97.22$\pm$0.19 \\
\midrule
UCI-HAR  & Tiny    & 2 & 64  & 108,597   & \bestall{89.20$\pm$3.32} \\
         & Default & 4 & 128 & 807,413   & \secondall{88.91$\pm$3.40} \\
         & Larger  & 6 & 128 & 1,201,141 & 86.94$\pm$2.06 \\
\bottomrule
\end{tabular}
\end{table}

Tiny has the highest mean accuracy on both datasets, exceeding Default by 0.26 points on Epilepsy and 0.29 points on UCI-HAR with about 14\% as many density-model parameters. Larger gives the lowest mean on both. These results show that a compact density model can supply useful per-timestep predictive statistics for patch classification. The small Tiny--Default differences relative to the reported seed variability leave their relative performance uncertain.

\subsection{Matched access to training data}
\label{app:har_matched_seed2026}

\paragraph{Three-seed comparison on UCI-HAR and Epilepsy.}

Within each dataset, seed, and label budget, every method receives the same labeled training examples. The 3 \texttt{ConceptTime} protocols use a common training configuration and compute budget. FullU fits the density model and vocabulary on all training inputs. LabelOnly fits both on the labeled subset. DensityFull/GMM-LabelOnly fits the density model on all inputs and the vocabulary on the labeled subset.

\begin{table}[!htbp]
\centering
\caption{Matched-label test accuracy (\%, mean $\pm$ standard deviation
across seeds 2021--2023). Column headings give the fraction of training
labels used.}
\label{tab:app_matched_three_seed}
\scriptsize
\setlength{\tabcolsep}{3pt}
\begin{tabular}{@{}lcccc@{}}
\toprule
& \multicolumn{2}{c}{UCI-HAR} & \multicolumn{2}{c}{Epilepsy} \\
\cmidrule(lr){2-3}\cmidrule(l){4-5}
Method & 1\% & 5\% & 1\% & 5\% \\
\midrule
CT-FullU                        & 65.73$\pm$0.45 & 68.86$\pm$0.31 & 96.64$\pm$0.41 & 96.96$\pm$0.24 \\
CT-LabelOnly                    & 69.16$\pm$2.44 & 67.33$\pm$2.61 & 96.22$\pm$0.46 & 97.36$\pm$0.09 \\
CT-DensityFull/GMM-LabelOnly    & 64.25$\pm$1.27 & 67.14$\pm$1.47 & 96.16$\pm$0.27 & 97.10$\pm$0.37 \\
\midrule
TS-TCC                          & 46.85$\pm$5.57 & 85.25$\pm$1.25 & 81.43$\pm$2.60 & 93.25$\pm$0.40 \\
TS-TCC-SSL                      & 72.51$\pm$4.46 & 80.55$\pm$1.72 & 94.28$\pm$0.93 & 95.30$\pm$0.17 \\
MiniRocket                      & 86.34$\pm$1.98 & 94.62$\pm$0.53 & 94.68$\pm$1.13 & 96.36$\pm$0.35 \\
TSLANet                         & 85.86$\pm$1.79 & 91.11$\pm$0.59 & 95.30$\pm$0.95 & 96.29$\pm$0.26 \\
\bottomrule
\end{tabular}
\end{table}

Across the 3 matched splits (Table~\ref{tab:app_matched_three_seed}), MiniRocket leads at both budgets on UCI-HAR. On Epilepsy, CT-FullU has the highest mean at 1\% labels and CT-LabelOnly at 5\%.

\paragraph{Additional UCI-HAR split.}

We repeated the UCI-HAR comparison on an additional split (seed 2026, Table~\ref{tab:app_har_matched_seed2026}). On this split, fitting only the vocabulary on the labeled subset while retaining full data access for the density model gives 77.13\% at 1\% labels. At 5\%, this protocol and FullU differ by 0.03 percentage points.

\begin{table}[!htbp]
\centering
\caption{UCI-HAR test accuracy (\%) on the additional matched split (seed 2026). Column headings give the fraction of training labels used.}
\label{tab:app_har_matched_seed2026}
\small
\begin{tabular}{@{}lcc@{}}
\toprule
Protocol & 1\% & 5\% \\
\midrule
CT-FullU & 74.63 & 79.88 \\
CT-LabelOnly & 75.25 & 76.09 \\
CT-DensityFull/GMM-LabelOnly & \textbf{77.13} & \textbf{79.91} \\
\bottomrule
\end{tabular}
\end{table}

\FloatBarrier

\section{Cross-Domain Transfer on Fault Diagnosis}
\label{app:fd_transfer}

\subsection{Transfer protocol}
\label{app:fd_transfer_protocol}

We distinguish the density model used to extract signatures from the vocabulary used to encode them. Let $V_s$ denote the source standardizer and fitted GMM. Given a density model $\mathcal M$ and a vocabulary $V$, write $Z(x;\mathcal M,V)$ for the sequence obtained by segmenting $x$, extracting patch signatures with $\mathcal M$, standardizing them with $V$, and computing their soft assignments. Panel~A of Table~\ref{tab:fd_transfer_maintext} holds $V_s$ fixed. Panel~B also evaluates target-fitted vocabularies. The transfer procedure supports label fractions of 5\% and 100\%; the controlled vocabulary comparison uses 5\%.

\begin{algorithm}[!htbp]
\caption{FD transfer with fixed or target-fitted concept vocabulary}
\label{alg:transfer_fd}
\footnotesize
\begin{algorithmic}[1]
\Require Source pipeline $(\mathcal M_s,V_s,g_s)$, target train/validation/test splits, label fractions $\mathcal R$
\For{each ordered pair $(s,t)$ with $s\ne t$}
  \State Evaluate $g_s(Z(x;\mathcal M_s,V_s))$ on target test examples (Direct).
  \State Train or load $\mathcal M_t$ fitted on unlabeled target-training inputs.
  \For{each target-label fraction $r\in\mathcal R$}
    \State Draw one stratified labeled subset of the target-training split for this seed.
    \For{each density/vocabulary configuration in the panel being evaluated}
      \State Select $\mathcal M\in\{\mathcal M_s,\mathcal M_t\}$ according to the configuration.
      \If{the configuration uses the source vocabulary}
        \State Set $V=V_s$.
      \Else
        \State Extract target-training signatures using the selected $\mathcal M$.
        \State Fit the standardizer and GMM to these signatures to obtain $V$.
      \EndIf
      \State Encode target splits with $Z(x;\mathcal M,V)$.
      \State Train the target classifier on the selected labeled subset and evaluate on the target test split.
    \EndFor
  \EndFor
\EndFor
\end{algorithmic}
\end{algorithm}

\subsection{Transfer accuracy by source operating condition}
\label{app:fd_transfer_groups}

We group the 12 directions from Panel~A of Table~\ref{tab:fd_transfer_maintext} by source condition. Direct transfer is more accurate on average for the 3 FD-D-source directions than for the 9 directions from FD-A/B/C (78.65\% versus 63.93\%). Training the classifier with 5\% of target labels brings the 2 groups to similar accuracies with a frozen source representation (90.45\% and 91.30\%, respectively). Using a target-trained density model gives a further 0.27-point gain for FD-D and a 4.28-point gain for FD-A/B/C. These results measure target classification accuracy. Appendix~\ref{app:fd_natural_ood} evaluates detection of these shifts.

\begin{table}[!htbp]
\centering
\caption{FD target test accuracy (\%) by source condition, aggregated from Panel~A of Table~\ref{tab:fd_transfer_maintext}. Entries give mean $\pm$ sample standard deviation across directions. Gains are percentage points relative to direct transfer within each row. Both few-shot settings use the source vocabulary.}
\label{tab:app_fd_transfer_groups}
\scriptsize
\setlength{\tabcolsep}{3pt}
\begin{tabular}{@{}lcccccc@{}}
\toprule
Source & Pairs & Direct & Source-$\mathcal M_\theta$ & Gain
& Target-$\mathcal M_\theta$ & Gain \\
\midrule
FD-A/B/C    & 9     &   63.93$\pm$26.58     &   91.30$\pm$7.52  &   +27.37    &   95.58$\pm$3.16  &   +31.65 \\
FD-D        & 3     &   78.65$\pm$19.06     &   90.45$\pm$1.72  &   +11.80    &   90.72$\pm$3.81  &   +12.08 \\
All         & 12    &   67.61$\pm$24.98     &   91.09$\pm$6.47  &   +23.48    &   94.37$\pm$3.84  &   +26.76 \\
\bottomrule
\end{tabular}
\end{table}

\FloatBarrier

\section{Fidelity, Selective Prediction, and Deletion Faithfulness}
\label{app:fidelity_evaluation}

\subsection{Detecting distribution shifts}
\label{app:fd_natural_ood}

\paragraph{Scoring and evaluation protocol.}

For OOD scoring, larger mean prototype distance indicates poorer coverage by the source vocabulary. For selective prediction on ID data, the confidence ordering has the opposite orientation: closer patches indicate greater coverage. The softmax OOD score $1-\max_y p(y\mid x)$ gives the same ranking as $-\max_y p(y\mid x)$.

\begin{algorithm}[!htbp]
\caption{OOD detection via concept fidelity}
\label{alg:ood_fidelity}
\footnotesize
\begin{algorithmic}[1]
\Require Frozen source pipeline, held-out ID inputs, OOD test inputs
\For{each ID or OOD test input $x$}
  \State Segment $x$ and extract standardized signatures $\phi_1,\ldots,\phi_K$.
  \State Compute $d_j=\min_m\|\phi_j-c_m\|_2$ and soft assignments $\pi_j$ for each patch.
  \State Obtain $p(y\mid x)$ from the frozen classifier on $Z=[\pi_1,\ldots,\pi_K]$.
  \State Set $s_{\mathrm{fid}}(x)=K^{-1}\sum_j d_j$ and $s_{\mathrm{sm}}(x)=1-\max_y p(y\mid x)$.
\EndFor
\If{evaluating synthetic corruptions}
  \State Use clean test inputs as ID and their corrupted versions as OOD.
  \State Compute AUROC for each corruption and severity. For TOTEM and VQShape, use their codebook distances and classifier confidence.
\Else
  \State Use held-out source inputs as ID and target-condition inputs as OOD.
  \State Compare the 6 scores in Table~\ref{tab:app_fd_natural_ood}; assign OOD the positive label.
  \State Report AUROC, AUPR-OOD, FPR95, and detection accuracy; select the accuracy threshold using source validation data only.
\EndIf
\end{algorithmic}
\end{algorithm}

\paragraph{Shifts across fault-diagnosis conditions.}

We complement the 6 synthetic corruption settings with all 12 ordered shifts between FD-A, FD-B, FD-C, and FD-D. We freeze each source pipeline and compare held-out source examples (ID) with examples from the target condition (OOD), using the protocol in Algorithm~\ref{alg:ood_fidelity}. Fidelity has the best mean on all 4 metrics (Table~\ref{tab:app_fd_natural_ood}). It improves mean AUROC from 0.528 for classifier confidence to 0.666. At 95\% OOD recall, however, its mean false-positive rate is still 0.719, and performance varies substantially across pairs.

\begin{table}[!htbp]
\centering
\caption{Detection of FD operating-condition shifts. Entries give mean $\pm$ standard deviation across the 12 directed pairs. AUPR-OOD treats target examples as positive. FPR95 is the false-positive rate at 95\% OOD recall.}
\label{tab:app_fd_natural_ood}
\scriptsize
\setlength{\tabcolsep}{4pt}
\begin{tabular}{@{}lcccc@{}}
\toprule
Score & AUROC $\uparrow$ & AUPR-OOD $\uparrow$
& FPR95 $\downarrow$ & Det.\ acc. $\uparrow$ \\
\midrule
Fidelity                & \textbf{0.666$\pm$0.187} & \textbf{0.674$\pm$0.186} & \textbf{0.719$\pm$0.343} & \textbf{0.605$\pm$0.145} \\
Mean density NLL            & 0.533$\pm$0.291 & 0.583$\pm$0.224 & 0.770$\pm$0.351 & 0.562$\pm$0.116 \\
Max density NLL             & 0.513$\pm$0.163 & 0.522$\pm$0.122 & 0.887$\pm$0.170 & 0.509$\pm$0.030 \\
$1-\max_y p(y\mid x)$       & 0.528$\pm$0.146 & 0.561$\pm$0.100 & 0.935$\pm$0.109 & 0.546$\pm$0.047 \\
Predictive entropy          & 0.517$\pm$0.152 & 0.560$\pm$0.108 & 0.922$\pm$0.122 & 0.546$\pm$0.047 \\
Mean predictive variance    & 0.522$\pm$0.255 & 0.543$\pm$0.160 & 0.787$\pm$0.312 & 0.505$\pm$0.019 \\
\bottomrule
\end{tabular}
\end{table}

\subsection{Selective prediction on Epilepsy}
\label{app:selective_epilepsy}

For a given confidence score, we retain the most confident examples and measure classification error as the retained fraction (coverage) increases. AURC is the area under this risk--coverage curve. Excess AURC (E-AURC) subtracts the area achieved by an optimal ordering of the same predictions, accounting for differences in base error.

\begin{table}[!htbp]
\centering
\caption{Selective prediction on Epilepsy. Lower values indicate better ranking of correct predictions ahead of errors. Entries give mean $\pm$ standard deviation across seeds 2021--2023. VQShape's deterministic logistic classifier gave identical predictions across these seeds. \capbest{red} marks the best value and \capsecond{yellow} the second best.}
\label{tab:app_selective_epilepsy}
\small
\setlength{\tabcolsep}{4pt}
\begin{tabular}{@{}llcc@{}}
\toprule
Method & Confidence score & AURC $\downarrow$ & E-AURC $\downarrow$ \\
\midrule
\texttt{ConceptTime} & Fidelity     &   \secondall{0.01060$\pm$0.00164}             &   \secondall{0.01032$\pm$0.00165} \\
& Mean prototype proximity          &   0.01062$\pm$0.00198            &   0.01034$\pm$0.00198 \\
& Max softmax                       &   \bestall{0.00144$\pm$0.00012}    &   \bestall{0.00117$\pm$0.00012} \\
\midrule
TOTEM & Mean codebook proximity     &   0.01514$\pm$0.00284             &   0.01319$\pm$0.00289 \\
& Max codebook proximity            &   0.01597$\pm$0.00381             &   0.01401$\pm$0.00386 \\
& Max softmax                       &   0.01547$\pm$0.00252             &   0.01351$\pm$0.00247 \\
\midrule
VQShape & Mean codebook proximity   &   0.11981                         &   0.11508 \\
& Max codebook proximity            &   0.10176                         &   0.09704 \\
& Max softmax                       &   0.02525                         &   0.02052 \\
\bottomrule
\end{tabular}
\end{table}

Fidelity and mean prototype proximity have nearly identical mean E-AURC (0.01032 and 0.01034). Both outperform all evaluated TOTEM and VQShape scores on this dataset (Table~\ref{tab:app_selective_epilepsy}). Within \texttt{ConceptTime}, max softmax gives the lowest E-AURC. The classifier sees only the concept sequence, so its confidence is a label-supervised readout of that bottleneck. Fidelity instead measures coverage by the learned prototypes and requires no additional confidence model or classifier weights.

\subsection{Patch-level deletion faithfulness}
\label{app:deletion_evaluation}

\subsubsection{Deletion protocol and attribution baselines}

All rankings are evaluated on the same frozen \texttt{ConceptTime} classifier. We progressively replace the highest-ranked patch assignments and measure how quickly accuracy falls. Fidelity ranks patches by their prototype-distance score. GradientSHAP uses classifier gradients, with attributions aggregated over concepts within each patch. For TimeX and TimeX++, we train timestep-mask generators to explain the frozen classifier's predictions and average each mask over the timesteps in each adaptive patch. Random and reversed fidelity rankings provide controls. The simplex-preserving comparison on Epilepsy evaluates fidelity, GradientSHAP, and random ordering.

\begin{algorithm}[!htbp]
\caption{Deletion-based faithfulness under a specified replacement}
\label{alg:deletion_faithfulness}
\footnotesize
\begin{algorithmic}[1]
\Require Frozen classifier $g$, test assignments $Z_i=[\pi_{i,1},\ldots,\pi_{i,K}]$, labels $y_i$, attribution method, replacement rule
\State Rank the $K$ patches of each example once using the attribution method.
\State Set $b_j=\mathbf{0}$ (original deletion), $b_j=\mathbf{1}/M$ (uniform), or $b_j=\overline{\pi}^{\,\mathrm{tr}}_j$ (mean training assignment at position $j$).
\For{$r=0,\ldots,K$}
  \For{each test example $i$}
    \State Copy $Z_i$ and replace each of its top-$r$ ranked assignments at position $j$ with $b_j$.
    \State Predict the label from the modified sequence with $g$.
  \EndFor
  \State Record test accuracy at deletion fraction $r/K$.
\EndFor
\State Integrate accuracy over the deletion fraction using the trapezoidal rule. Lower area indicates an earlier loss of classification accuracy.
\end{algorithmic}
\end{algorithm}

\subsubsection{Deletion with valid concept assignments}
\label{app:simplex_deletion}

A zero vector has no probability mass, whereas each observed patch assignment sums to one. We therefore repeat the deletion experiment with 2 replacements that lie on the probability simplex: a uniform assignment and the mean training assignment at the same patch position (Algorithm~\ref{alg:deletion_faithfulness}). On Epilepsy, across 3 seeds, both fidelity and GradientSHAP give lower deletion AUC than random ordering under both replacements. With uniform replacement, fidelity gives the lowest AUC (0.7514), followed by GradientSHAP (0.7634) and random ordering (0.7879). These comparisons measure the importance of patches to the frozen classifier under the specified replacement.

\FloatBarrier

\section{Qualitative Examples}
\label{app:vizz}

\subsection{Epilepsy recordings}

\begin{figure}[!htbp]
    \centering
    \includegraphics[width=1\linewidth]{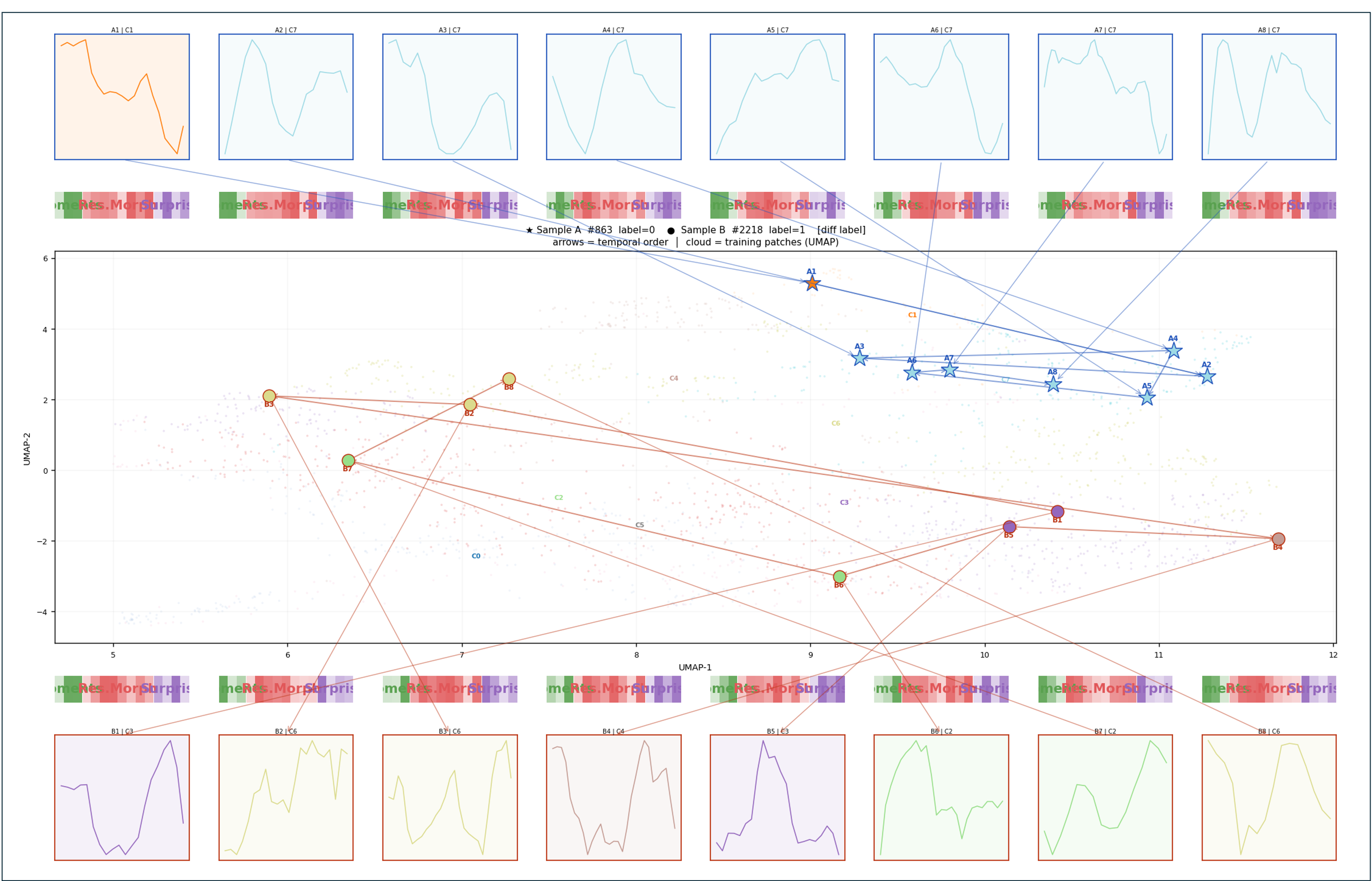}
    \caption{Seizure and non-seizure recordings in concept space.}
    \label{fig:placeholder1}
\end{figure}

\begin{figure}[!htbp]
    \centering
    \includegraphics[width=1\linewidth]{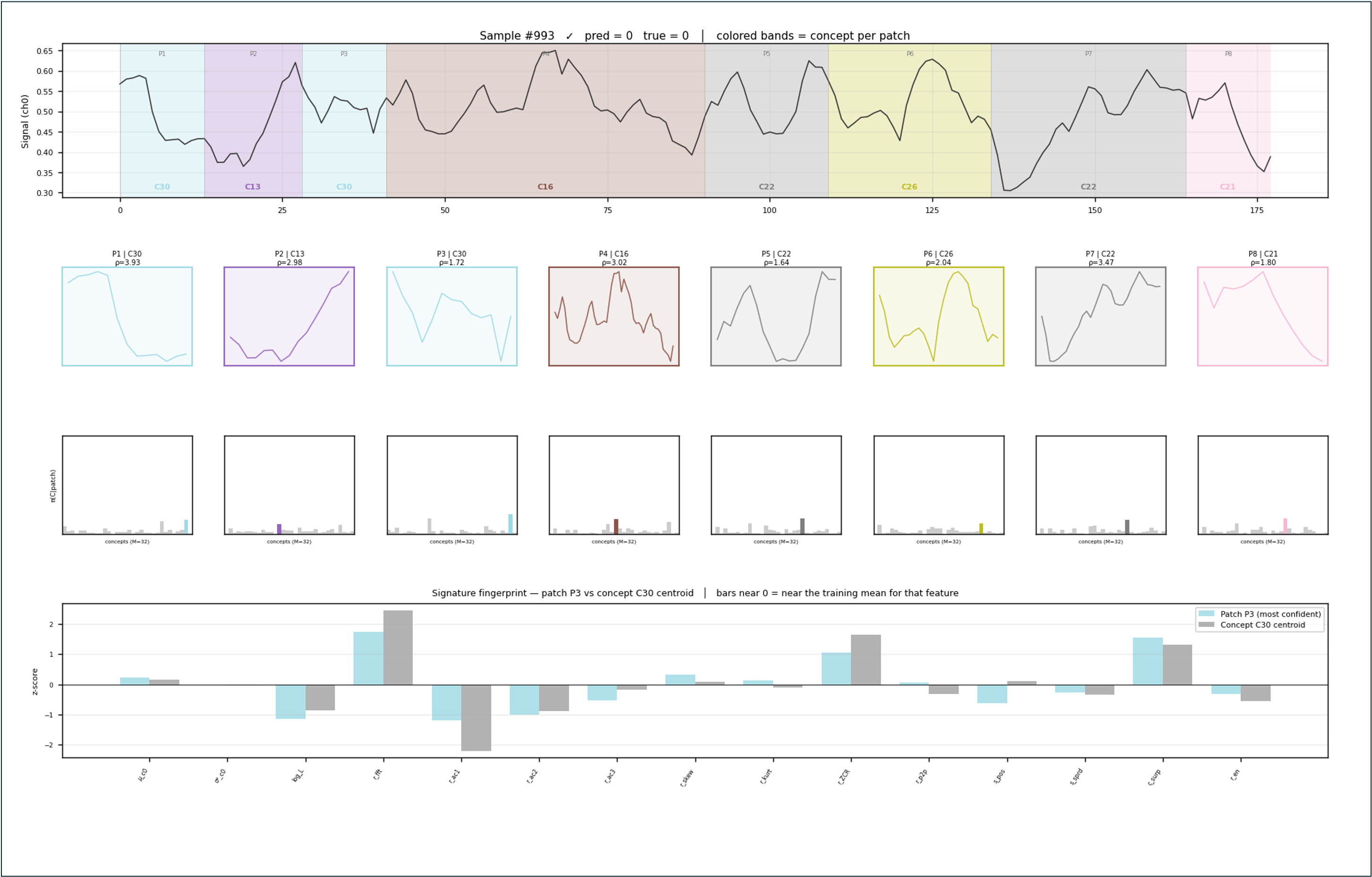}
    \caption{Concept assignment confidence across segments of an Epilepsy recording.}
    \label{fig:placeholder2}
\end{figure}

\begin{figure}[!htbp]
    \centering
    \includegraphics[width=1\linewidth]{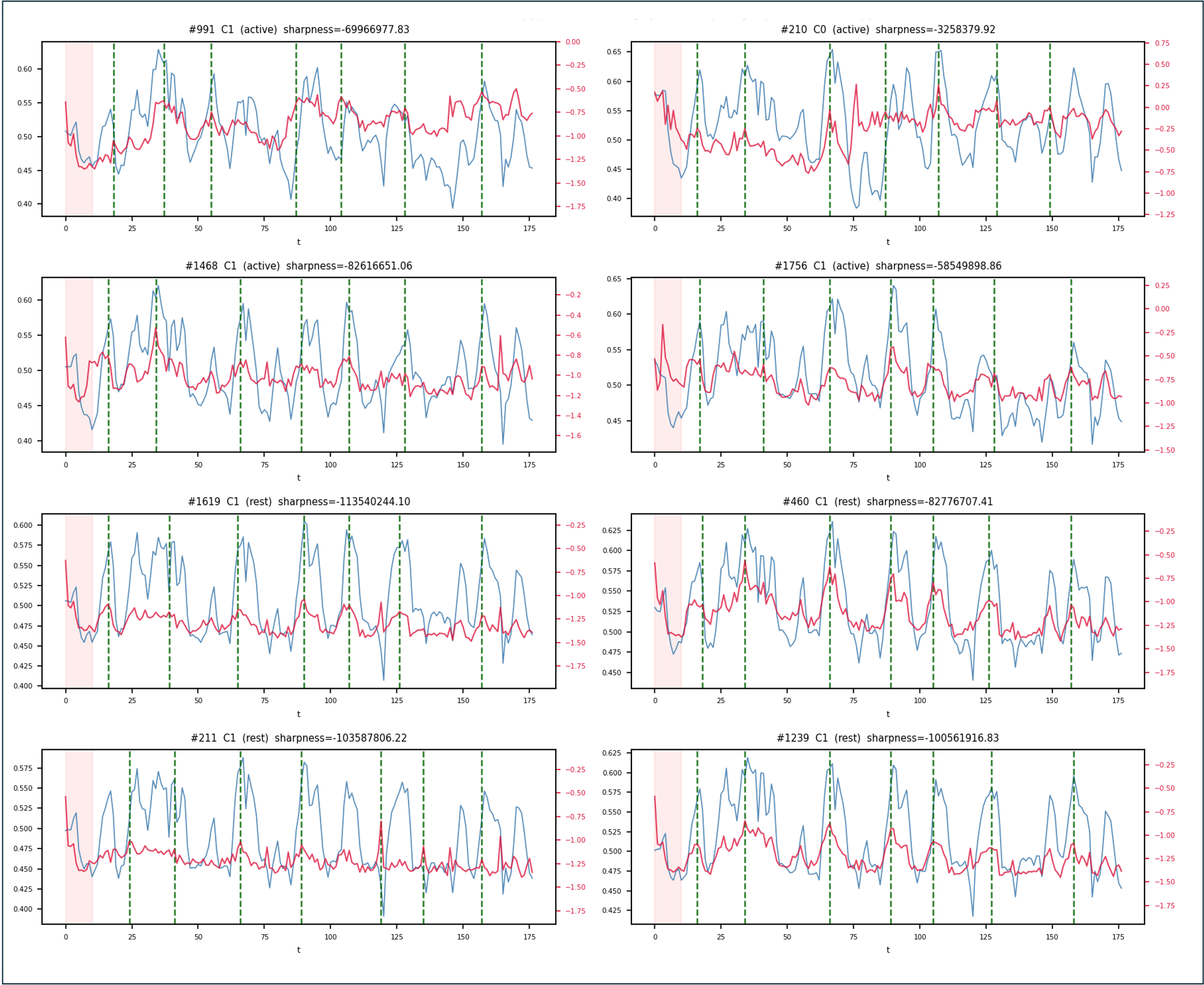}
    \caption{Segment boundaries overlaid on Epilepsy recordings.}
    \label{fig:placeholder3}
\end{figure}

\begin{figure}[!htbp]
    \centering
    \includegraphics[width=1\linewidth]{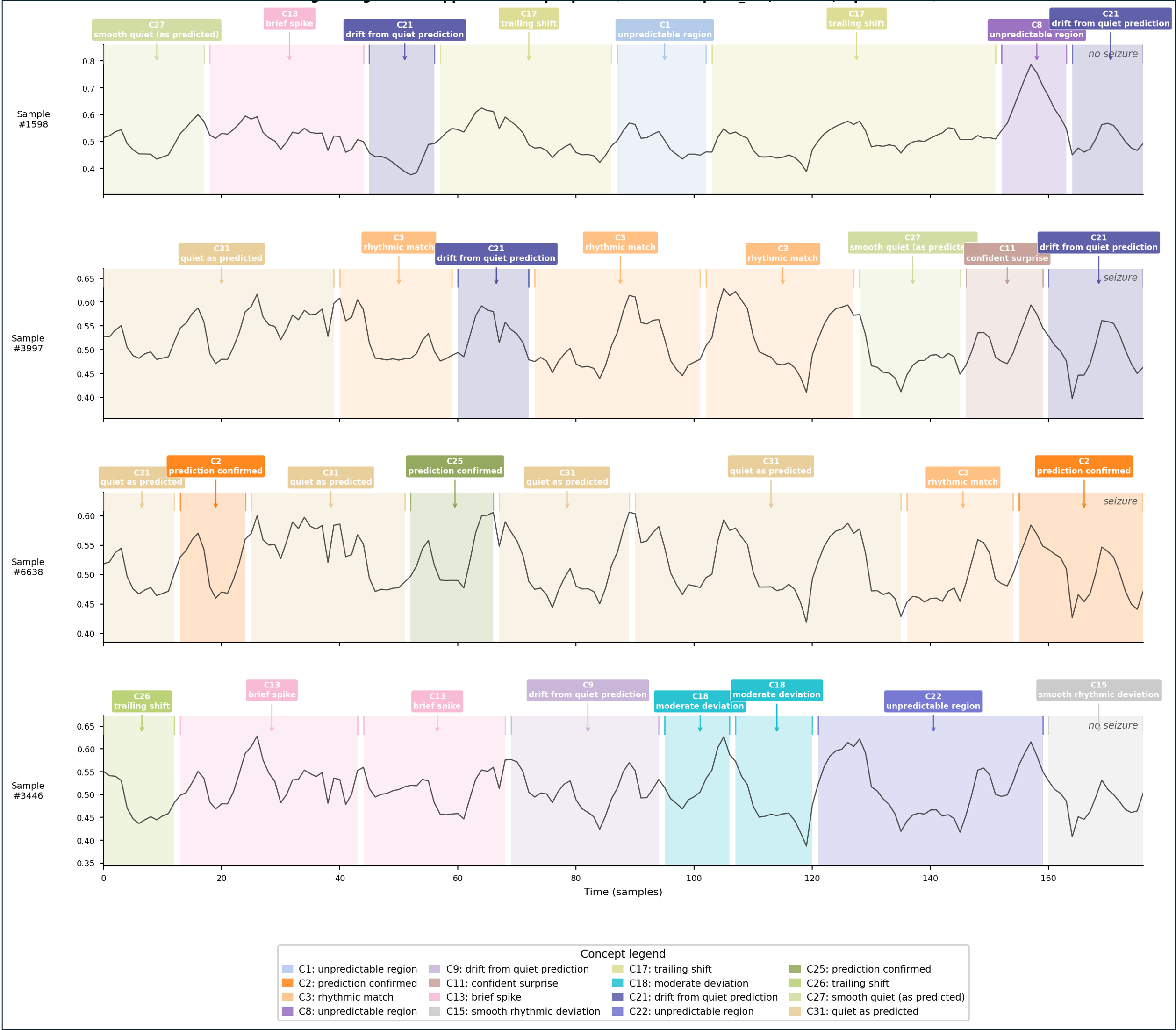}
    \caption{Examples of adaptive segments in Epilepsy recordings.}
    \label{fig:placeholder4}
\end{figure}

\begin{figure}[!htbp]
    \centering
    \includegraphics[width=1\linewidth]{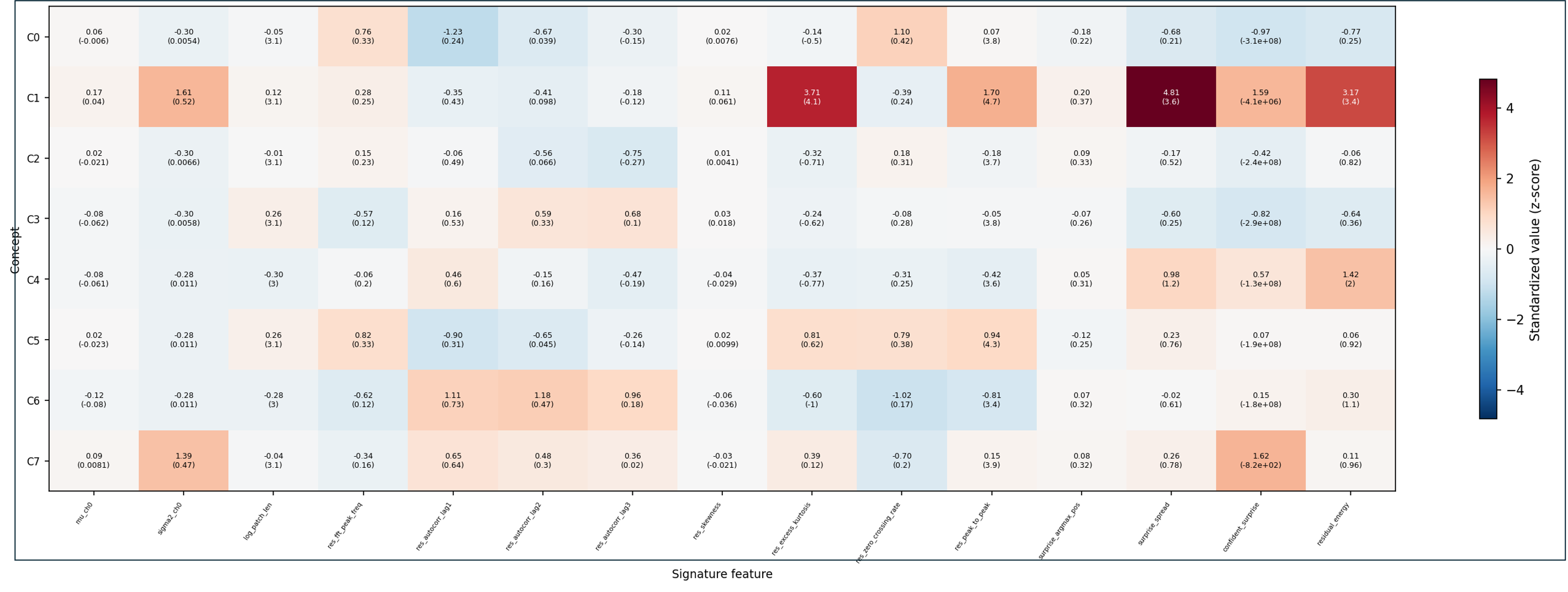}
    \caption{Signature features of the learned Epilepsy concepts.}
    \label{fig:placeholder6}
\end{figure}

Figures~\ref{fig:placeholder1} and~\ref{fig:placeholder2} show how recordings map to concepts. Figures~\ref{fig:placeholder3} and~\ref{fig:placeholder4} connect those assignments to time intervals in the original signals. Figure~\ref{fig:placeholder6} shows the signature features associated with the concepts.

\FloatBarrier

\subsection{Human activity recordings}

\begin{figure}[!htbp]
    \centering
    \includegraphics[width=1\linewidth]{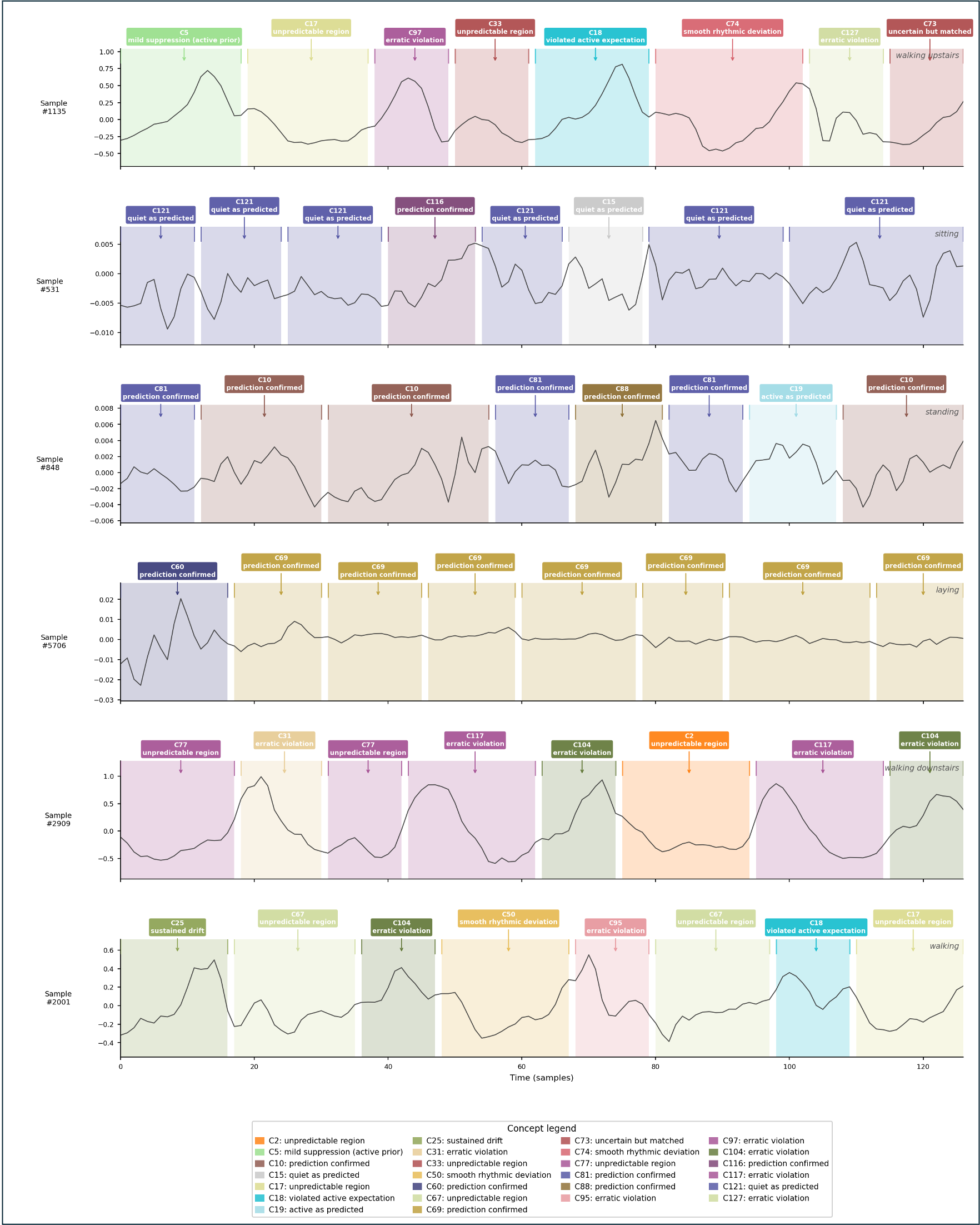}
    \caption{Examples of adaptive segments in UCI-HAR recordings.}
    \label{fig:placeholder5}
\end{figure}

\begin{figure}[!htbp]
    \centering
    \includegraphics[width=1\linewidth]{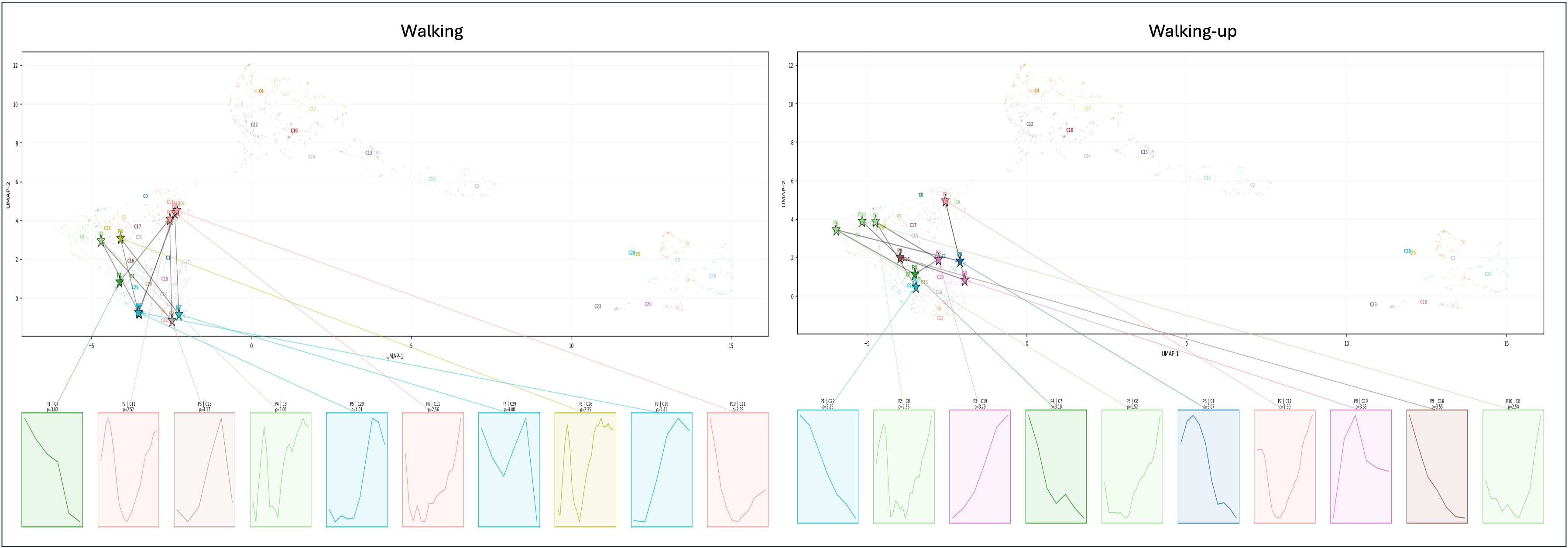}
    \caption{Concept assignments for walking and walking upstairs on UCI-HAR.}
    \label{fig:placeholder8}
\end{figure}

Figure~\ref{fig:placeholder5} shows segment assignments on UCI-HAR. Figure~\ref{fig:placeholder8} compares walking and walking upstairs in the learned concept space.

\FloatBarrier

\subsection{Fidelity and abstention}

\begin{figure}[!htbp]
    \centering
    \includegraphics[width=1\linewidth]{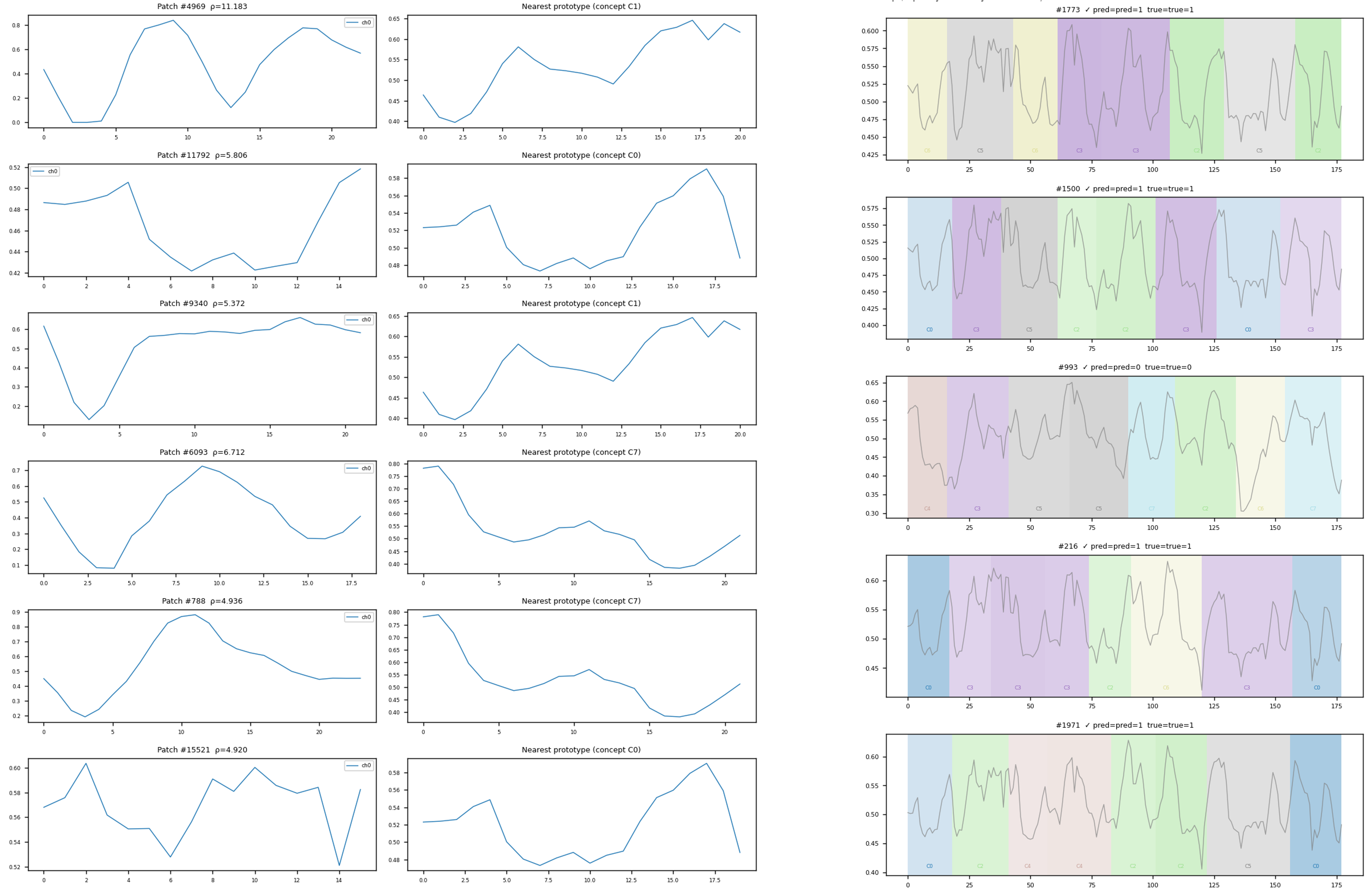}
    \caption{Abstention decisions (left) and fidelity-weighted segment-to-concept assignments (right).}
    \label{fig:placeholder7}
\end{figure}

Figure~\ref{fig:placeholder7} illustrates the abstention decision alongside the segment-level assignments used to inspect a recording.

\section*{Use of Large Language Models}
We refined parts of this paper using a large language model (ChatGPT, GPT-5), exclusively for language polishing and improving clarity of exposition. The authors conceived and executed all ideas, methodology, experiments, and analyses.

\section*{Broader Impacts}
In healthcare and industrial monitoring, \texttt{ConceptTime} can help users inspect a prediction by tracing it to specific signal segments and comparing those segments with learned prototypes. A low fidelity score can flag recordings that the learned vocabulary describes poorly and help prioritize them for human review.

A familiar signal pattern can still be misclassified, and a learned prototype may reflect a sensor artifact as readily as a physiological event or machine fault. Treating high fidelity as evidence of correctness could therefore create false reassurance. Before deployment, users should validate predictions and abstention thresholds on data from the intended population, sensors, and operating conditions, and retain human oversight for consequential decisions.

\end{document}